\documentclass[journal]{IEEEtran}

\usepackage[american]{babel}

\usepackage{graphicx}
\graphicspath{{figure/}{voc12-coco14/}{photo/}}
\usepackage[caption=false,position=bottom]{subfig}
\usepackage{amsmath,amssymb}
\usepackage[bold]{hhtensor}
\usepackage{multirow} 
\usepackage[breaklinks=true,colorlinks,bookmarks=false]{hyperref}

\newcommand{\squishlist}{
 \begin{list}{$\bullet$}
  { \setlength{\itemsep}{0pt}
     \setlength{\parsep}{1pt}
     \setlength{\topsep}{1pt}
     \setlength{\partopsep}{0pt}
     \setlength{\leftmargin}{1.5em}
     \setlength{\labelwidth}{1em}
     \setlength{\labelsep}{0.5em} } }
\newcommand{\squishend}{
  \end{list}  }

\DeclareRobustCommand\onedot{\@onedot}
\def\@onedot{.}
\def\eg{\emph{e.g}\onedot} 
\def\ie{\emph{i.e}\onedot}

\def\etc{\emph{etc}\onedot}

\begin{document}
\title{Learning to Discover Multi-Class Attentional Regions for Multi-Label Image Recognition}
\author{Bin-Bin Gao and Hong-Yu Zhou
\thanks{Bin Bin Gao is with the Tencent YouTu Lab,  Shenzhen 518057, China. (E-mail: {gaobb}@lamda.nju.edu.cn. Corresponding author: Bin-Bin Gao)}
\thanks{Hong-Yu Zhou is with the Department of Computer Science, The University of Hong Kong, Hong Kong. (E-mail: whuzhouhongyu@gmail.com)}
}

\markboth{Accepted by IEEE Trans. Image Processing}{Accepted by IEEE TIP}

\maketitle

\begin{abstract}
Multi-label image recognition is a practical and challenging task compared to single-label image classification. However, previous works may be suboptimal because of a great number of object proposals or complex attentional region generation modules. In this paper, we propose a simple but efficient two-stream framework to recognize multi-category objects from global image to local regions, similar to how human beings perceive objects. To bridge the gap between global and local streams, we propose a multi-class attentional region module which aims to make the number of attentional regions as small as possible and keep the diversity of these regions as high as possible. Our method can efficiently and effectively recognize multi-class objects with an affordable computation cost and a parameter-free region localization module. Over three benchmarks on multi-label image classification, our method achieves new state-of-the-art results with a single model only using image semantics without label dependency. In addition, the effectiveness of the proposed method is extensively demonstrated under different factors such as global pooling strategy, input size and network architecture. Code has been made available at~\url{https://github.com/gaobb/MCAR}.
\end{abstract}

\IEEEpeerreviewmaketitle

\section{Introduction}
\IEEEPARstart{C}{onvolutional} Neural Networks (CNNs) have made revolutionary breakthroughs on various computer vision tasks. For example, single-label image recognition~(SLR), as a fundamental vision task, has surpassed human-level performance~\cite{he2015delving} on large-scale ImageNet. Unlike SLR, multi-label image recognition~(MLR) needs to predict a set of objects or attributes of interest present in a given image. Meanwhile, these objects or attributes usually have complex variations like spatial location, object scale and occlusion~\etc. Nonetheless, MLR still has wide applications such as scene understanding~\cite{shao2015deeply}, face or human attribute recognition~\cite{liu2015deep,li2016human} and multi-object perception~\cite{wei2015hcp} \etc. These make MLR become a practical and challenging task. In recent years, a significant amount of learning approaches have been proposed to dealing with multi-label data~\cite{zhang2013review}.

MLR can be simply addressed by using SLR framework to predict whether each category object presents or not. Recently, there are many works using deep CNNs to improve the performance of MLR. These works can be roughly divided into three types: spatial information~\cite{wei2015hcp,yang2016exploit}, visual attention~\cite{chen2018recurrent,wang2017multi,zhu2017learning,guo2019visual} and label dependency~\cite{wang2016cnn,chen2018order,chen2019multi,chenlearning}. 

Since the goal of MLR is to predict a set of object categories instead of producing accurate spatial locations of all possible objects, we argue that it is not necessary to waste computation resource for hundreds of object proposals in HCP~\cite{wei2015hcp} or consume labor cost for the bounding box annotation of objects in Fev+Lv~\cite{yang2016exploit}. RARL~\cite{chen2018recurrent} and RDAL~\cite{wang2017multi} introduce a reinforcement learning module and a spatial transformer layer to localize attentional regions, respectively, and sequentially predict label distribution based on generated regions. The main problem of these two methods is that the generated attentional regions are always category-agnostic and it is also difficult to guarantee the diversity of these local regions. In fact, we should ask the number of attentional regions to be as small as possible while maintaining the high diversity. Recently, MLGCN~\cite{chen2019multi,chen2021learning} and SSGRL~\cite{chenlearning} try to model the label dependency with graph CNN to boost the performance of MLR. However, in this paper, we aim to improve the performance of MLR with only image semantics.

In order to exploit the semantic information of image, let us recall how we humans recognize multiple objects appeared in an image. Firstly, people may have a glimpse of a given image to discover some possible object regions from a global view. Then, these possible object regions guide the eye movements and help to make decisions on specific object categories following a region-by-region manner. In other words, most of time we humans difficultly recognize multi-objects using a single glance but at least two steps from a global view to local regions. In fact, there actually exists evidence in cognitive science that global visual processing precedes local reaction in visual perception~\cite{navon1969forest}. Also, such global-to-local mechanism is supported by studies in neurobiology~\cite{hegde2008time} and psychology~\cite{flevaris2014attending}. In this paper, we wonder if machines can acquire the learning ability like humans to recognize multi-objects.

\begin{figure*}[t]
 \centering
 {\includegraphics[width= 0.9\textwidth]{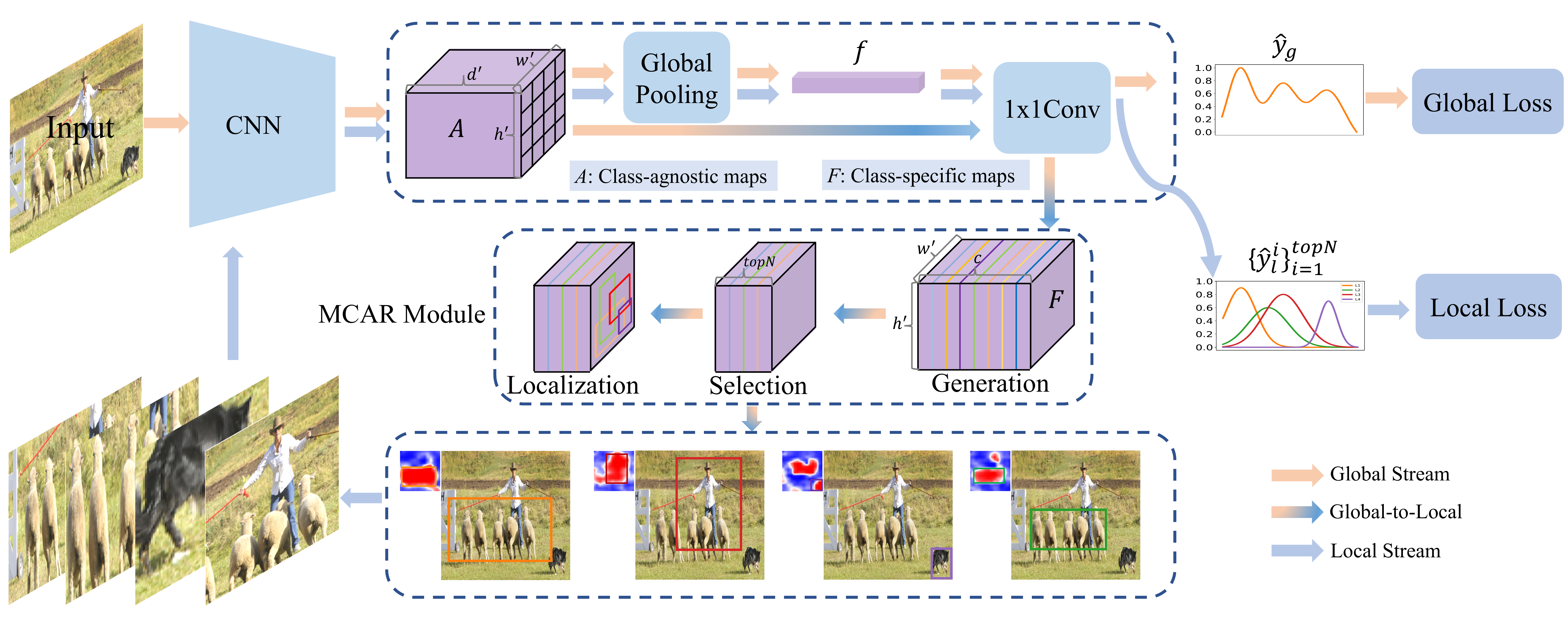}}
\caption{The pipeline of our MCAR framework for multi-label image recognition. MCAR firstly feeds an input image into a deep CNN model to extract its global feature representation through the global image stream. Then, the multi-class attentional region module roughly localizes possible object regions by integrating that information from the global stream. Finally, these localized regions are fed to the shared CNN to obtain their predicted class distributions through the local region stream. At the inference stage, MCAR aggregates predictions from global and local streams with category-wise max-pooling and produces the final prediction.}\label{fig:pipeline}
\end{figure*}

Inspired by this observation, we propose a novel multi-label image recognition framework with Multi-Class Attentional Regions~(MCAR) as illustrated in Fig.~\ref{fig:pipeline}. This framework contains a global image stream, a local region stream, and a multi-class attentional region module. Firstly, the global image stream takes an image as the input for a deep CNN and learns global representations supervised by the corresponding labels. Then, the multi-class attentional region module is used to discover possible object regions with the information from the global stream, which is similar to the way we recognize multiple objects. Finally, these localized regions are fed to the~\emph{shared} CNN to obtain their predicted class distributions using the local region stream. The local region stream can recognize objects better since it flexibly focuses on details of each object which helps to alleviate the difficulty of recognition for these objects at different spatial locations and object scales.
 
The contributions of this paper can be summarized as follows. 
\squishlist 
\item Firstly, we present a multi-label image recognition framework that can efficiently and effectively recognize multi-objects following a global to local manner. To the best of our knowledge, the learning mechanism of global to local in a unified model is the first time being proposed to find possible regions for multi-label images.

\item Secondly, we propose a simple but effective multi-class attentional region module which includes three steps: generation, selection, and localization. In practice, it can dynamically generate a small number of attentional regions while keeping their diversity as high as possible.

\item Thirdly, we achieve new state-of-the-art results on three widely used benchmarks with only a single model. Our method provides an affordable computation cost and needs no extra parameters.

\item In addition, we also extensively demonstrate the effectiveness of the proposed method under different conditions like global pooling strategy, input size and network architecture.
\squishend

The rest of this paper is organized as follows. We first review the related work in Section~\ref{rws}. 
Then, Section~\ref{mcarf} proposes our approach, including two-stream framework, MCAR module (from global to local) and two-stream learning. After that, the experiments are reported in Section~\ref{exps}. Finally, Section~\ref{discuss} presents discussions and the conclusion is given in Section~\ref{cons}.

\section{Related Works}\label{rws}
Recently, many efforts have been devoted into multi-label image recognition, using spatial information~\cite{wei2015hcp,yang2016exploit}, visual attention~\cite{chen2018recurrent,wang2017multi,zhu2017learning,guo2019visual} and label dependency~\cite{wang2016cnn,chen2018order,chen2019multi,chenlearning}. In this section, we briefly
review these related approaches.

\noindent \textbf{Spatial Information.} 
How to utilize the spatial information of image is very crucial for almost all visual recognition tasks such as image recognition~\cite{lazebnik2006beyond,he2015spatial}, object detection~\cite{girshick2014rich} and semantic segmentation~\cite{zhao2017pyramid,chen2017rethinking}. It is closely related to how to design (or learn) effective features. The reason is that objects usually present with different scales at different spatial locations. HCP~\cite{wei2015hcp} uses EdgeBox~\cite{zitnick2014edge} or BING~\cite{cheng2014bing} to generate hundreds of object proposals for each image using a like RCNN~\cite{girshick2014rich} method, and aggregates prediction scores of these proposals to obtain the final prediction.  However, a large number of proposals usually bring a huge computation cost. Fev+Lv~\cite{yang2016exploit} generates proposals using bounding box annotations. Their approach combined the local proposal features and global CNN features to produce the final feature representations. It reduces the number of proposals but introduces the labor cost of annotation.

\noindent \textbf{Visual Attention.}  
Attention mechanism has been widely used in many vision tasks, such as visual tracking~\cite{bazzani2011learning}, fine-grained image recognition~\cite{fu2017look}, image captioning~\cite{xu2015show}, image question answering~\cite{anderson2018bottom}, and semantic segmentation~\cite{hong2016learning}. RARL~\cite{chen2018recurrent}  uses a recurrent attention reinforcement learning module~\cite{mnih2014recurrent} to localize a sequence of attention regions and further predict label scores conditioned on these regions. Instead of reinforcement learning in RARL, RDAL~\cite{wang2017multi} introduces a spatial transformer layer~\cite{jaderberg2015spatial,yu2019delta} for localizing attentional regions from an image and an LSTM unit to sequentially predict the category distribution based on features of these localized regions. Unlike RARL and RDAL, SRN~\cite{zhu2017learning} and  ACfs~\cite{guo2019visual} combine attention regularization loss and multi-label loss to improve performance. Specifically, SRN~\cite{zhu2017learning} captures both spatial semantic and label correlations based on the weighted attention map, while ACfs~\cite{guo2019visual} enforces the network to learn attention consistency that the classification attention map should follow the same transformation when input image is spatially transformed.

\noindent \textbf{Label Dependency.} 
In order to exploit label dependency, CNN-RNN~\cite{wang2016cnn} jointly learns image feature and label correlation in a unified framework composed of a CNN module and an LSTM layer. The limitation is that it requires a pre-defined label order for model training. Similar to~\cite{wang2016cnn},~\cite{xu2020joint} also jointly learn multi-label classifiers with both spatial object relationships and semantic label correlations. 
Order-Free RNN~\cite{chen2018order} relaxes the label order constraint via learning visual attention model and a confidence-ranked LSTM. But it requires an explicit module for removing duplicate prediction labels and needs a threshold for stopping the sequence outputs.  In order to alleviate the issues, PLA~\cite{yazici2020orderless} proposes two alternative losses which dynamically order the labels based on the prediction label sequence of an LSTM model. Recently, SSGRL~\cite{chenlearning} directly uses a graph convolutional network to model the label dependency among all labels. 

There have been some other attempts on multi-label researches, such as multi-label image retrieval~\cite{7438833}, multi-label dictionary learning~\cite{jing2016multi}, zero-shot~\cite{ji2020deep,ji2020deep} and few-shot~\cite{9207855} multi-label classification. While in this paper, we deliberately avoid using any information from label dependency and aim to improve the performance of multi-label recognition with only image semantic information. We leave them as future works to further boost recognition performance or extend application fields by integrating the label correlation and other paradigms to our framework.

\section{MCAR Framework}\label{mcarf}
In this section, we firstly present a two-stream framework which contains a global image stream and a local region stream. Then, we elaborate the multi-class attentional region module, which tries to bridge the gap between global and local views. Finally, we present the optimization details of our framework.

\subsection{Two-Stream Framework}
\noindent \textbf{Global Image Stream.}  
Given an input image $I \in \mathbb{R}^{h\times w\times 3}$, where $h$, $w$ are the image's height and width. Let's denote its corresponding label as $\vec y={[y^1, y^2, \cdots, y^C]}^T$, where $y^i$ is a binary indicator. $y^i=1$ if image $I$ is tagged with label $i$, otherwise $y^i=0$. $C$ is the number of all possible categories in this dataset.
 
We assume that $A ={\mathcal F} (I; \vec \theta)$ is the activation map of the last convolutional layer of a CNN, where $\theta$ denotes the parameters of the CNN and $A \in \mathbb{R}^{h^\prime \times w^\prime \times d^\prime}$. Then, a global pooling function $\mathcal {P}(\cdot)$ encodes the activation map $A$ to a single vector $\vec f \in \mathbb{R}^{ 1 \times 1\times d^\prime}$, \ie, $ \vec f = \mathcal {P}(A) $. Here $\vec f$ can be considered as a global feature representation of the image $I$. In order to get it's prediction score, a 1$\times$1 fully convolutional layer transfers $\vec f$ to $\vec x \in \mathbb{R}^{C}$ by
 \begin{equation}\label{eq:linear}
 \vec x =  W^T\vec f +  \vec b.
 \end{equation}
 We then use a sigmoid function $\sigma(\cdot)$ to turn $\vec x$ into a range $[0,1]$, that is 
 \begin{equation}\label{eq:sigmoid}
 \vec {\hat {y}_g} = \frac {1}{1+\exp(-\vec x)},
 \end{equation}
 where $\vec {\hat{y}_g}$ stands for the global prediction distribution.
 
\noindent \textbf{Local Regions Stream.} 
Local stream is, in fact, to perform a multi-instance multi-label learning~\cite{zhou2012multi}. By decomposing an image into object regions, each image becomes a bag containing several positive instances, \ie, regions
with the target objects, and negative instances, \ie, regions with background or other objects. We assume that $\{L_1, L_2, \cdots, L_N\}$ is a set of $N$ local regions cropped from input image $I$. These local regions are firstly resized to the input size by bilinear upsampling. Then, they are fed to the shared CNN (with the global stream) to get prediction distributions $\{\vec {\hat {y}_{L_1}}, \vec {\hat {y}_{L_2}}, \cdots, \vec {\hat {y}_{L_N}}\}$ with Eq.~\ref{eq:linear} and~\ref{eq:sigmoid}. Finally, these local region distributions are aggregated by a category-wise max-pooling operation:
 \begin{equation}\label{eq:classmax}
 \hat {y}_l^i = \max \big(\hat y_{L_1}^i , \hat y_{L_2}^i , \cdots, \hat y_{L_N}^i \big),
 \end{equation}
where $\hat {y}_l^i$ is the $i$-th category score of the local prediction $\vec {\hat {y}_l}$. The subscript $l$ means the distribution is from $N$ local regions.

Note that \emph{the global and local streams share the same network without introducing additional parameters}. It is obviously different from the classical two-stream architecture which usually contains two parallel subnetworks. The inputs of our two-stream are the whole image and local regions from it, respectively. These local regions are dynamically generated by using the information of the global stream. Therefore, it is also different from the existing methods whose inputs are always two parallel views like video frame and optical flow in video classification~\cite{Simonyan14}.

During the training stage, we jointly train these two streams. At the early stage of learning, there may be little difference between the number of positive and negative instances (local regions). With the gradual convergence of the global stream, positive instances will dominate the local stream and thus also tend to converge.  At the inference stage, we fuse the predictions from global stream ($\vec {\hat {y}_g}$) and local stream ($\vec {\hat {y}_l}$) with a category-wise max-pooling operation to generate the final predicted distribution of image $I$.

\subsection{From Global to Local}
Potential object regions are not available in image-level labels, which must be generated in an efficient manner. The desirable generation module and candidate regions should satisfy some basic principles. First, the diversity of candidate regions should be as high as possible such that they can cover all possible objects of a given multi-label image. Second, the number of these candidate regions should be as small as possible in order to ensure efficiency. In contrast, more candidate regions require more computation resources since these regions need to be fed to the shared CNN simultaneously. Last but not least, the candidate regions generation module should have a simple network architecture and few parameters to alleviate the computation cost and storage overhead.

\noindent \textbf{Attentional Maps Generation.} 
The class activation mapping method~\cite{zhou2016learning} intuitively shows the discriminative image regions and helps us understand how to identify a particular category with a CNN.  To obtain class-specific activation maps, we directly apply the 1$\times$1 convolutional layer to the class-agnostic activation maps $A$ from the global stream, that is
 \begin{equation}\label{eq:cam}
 F = W^T A +  \vec b,
 \end{equation}
 where $F \in \mathbb{R}^{h^\prime \times w^\prime \times c}$. The class-specific activation map of the $i$-th category is denoted as $F^{i}\in  \mathbb{R}^{h^\prime \times w^\prime}$ and it directly indicates the importance of the activation map at spatial leading to the classification of an image to class $i$. 
  
The discriminative class regions of a specific $F^{i}$ are significantly different among all possible class maps $\{F^i\}_{i=1}^C$. If we employ class maps  $\{F^i\}_{i=1}^C$ to localize the potential object regions then it is easy to satisfy the first principle: to increase the diversity of different proposals.
 
\noindent \textbf{Attentional Maps Selection.} 
The number of class activation maps is equal to that of all categories associated with a dataset. For example, there are 20 and 80 categories on PASCAL VOC and MS-COCO datasets, respectively. If we use all class maps, it leads to two problems. First, the generated regions are too many to ensure efficiency. Second, a majority of regions will be redundant or meaningless because an image usually consists of a few instances.

\begin{figure}[t]
 \centering
 {\includegraphics[width= 0.6\columnwidth]{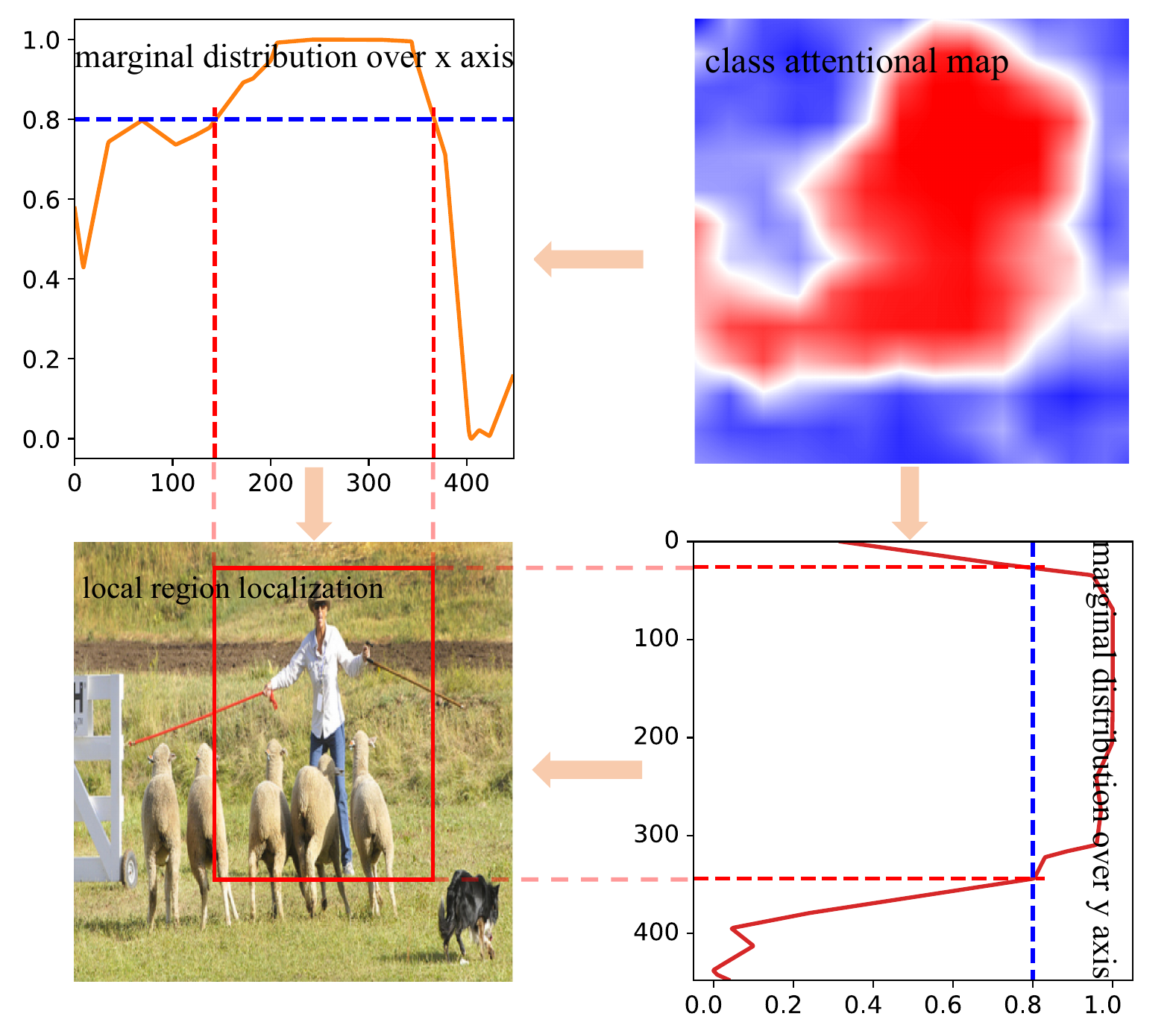}}
\caption{The visualization of local region localization with class attentional map. We firstly decompose the class attentional map into two marginal distributions along row and column. Then, the class attentional region is localized by these two marginal distributions.}\label{fig:mcar}
\end{figure}

A fact is that the predicted distribution will be close to the ground-truth distribution with the learning of the network which is supervised by ground-truth labels. It is a reasonable assumption that the high category confidence means that the corresponding object presents on the image with a high probability. Therefore, we sort the predicted scores $\vec {\hat {y}_g}$ (whose dimension is equal to the number of classes) following a descending order and select the $topN$ class attentional maps. In experiments, we can see that a satisfied performance can be achieved when the $topN$ is a small number (such as 2 or 4) which is \emph{far less than the number of all categories}. Another benefit is that the proposed method may force network to implicitly learn label correlation if selective attentional maps don't fully cover all object categories. This is because the local stream is also supervised by the ground-truth label distribution.

\noindent \textbf{Local Regions Localization.} 
We still denote $topN$ class attentional maps as $\{F^i\}_{i=1}^{topN}$ for notation simplification. Each $F^i$ is normalized to the range $[0,1]$ by a sigmoid function~(Eq.~\ref{eq:sigmoid}). Furthermore, we simply upsample  $F^i$ to the input size to align the spatial semantics between $F^i$ and the input image $I$.

\begin{figure}[t]
 \centering
 \subfloat[single peak]
 {\includegraphics[width= 0.28\columnwidth]{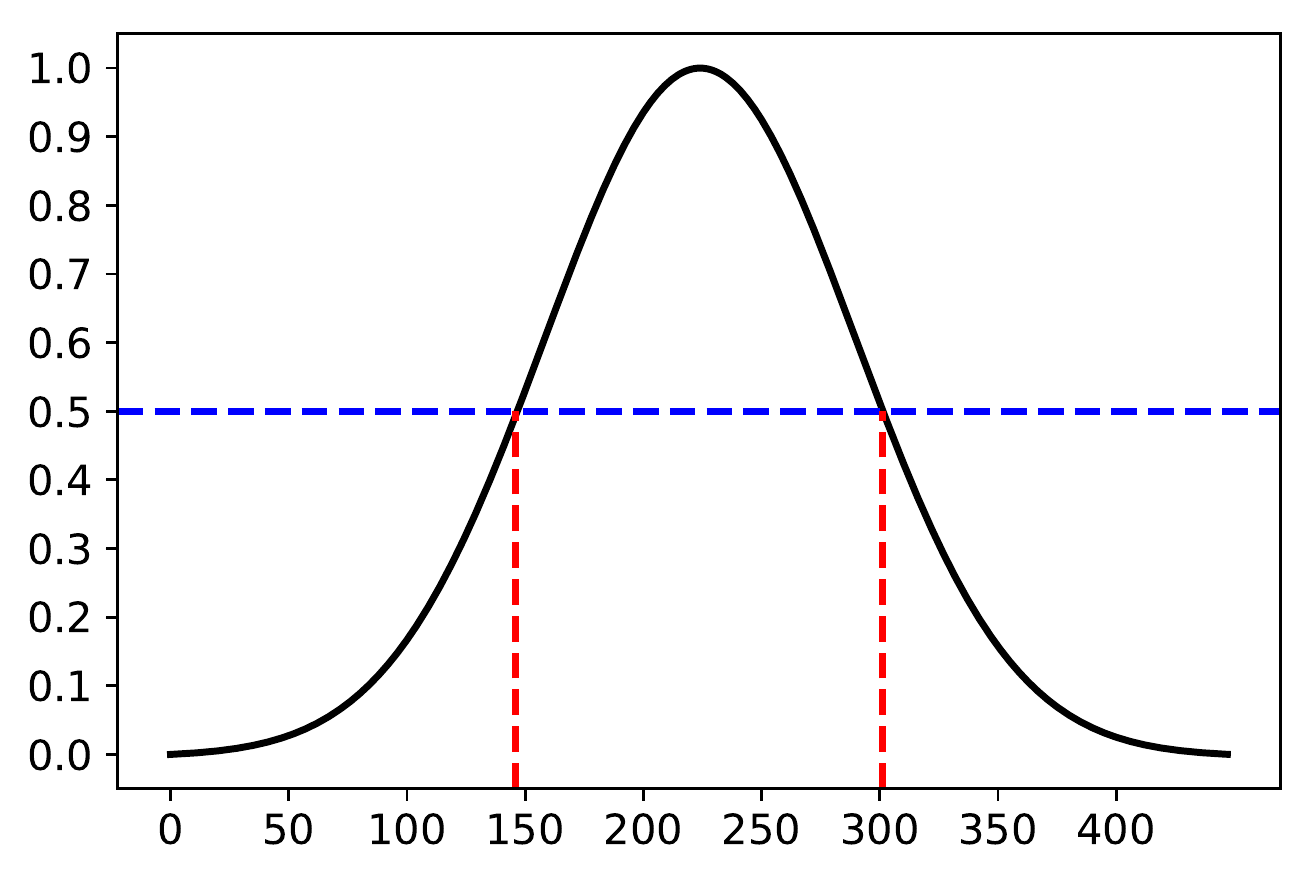}\label{fig:loceg1}} \hspace{15pt}
 \subfloat[multiple peaks]
 {\includegraphics[width= 0.28\columnwidth]{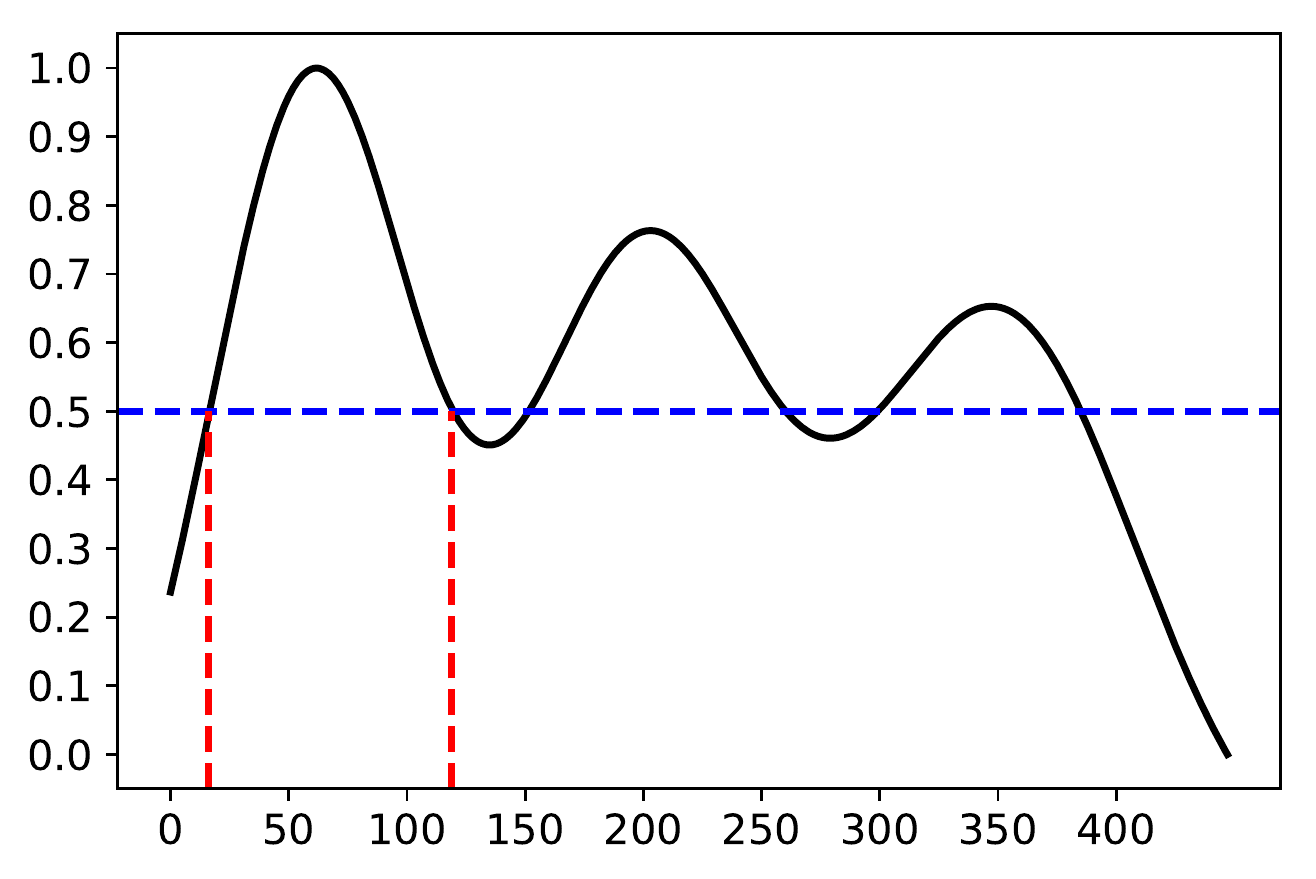}\label{fig:loceg2}}\hspace{15pt}
  \subfloat[multiple peaks]
 {\includegraphics[width= 0.28\columnwidth]{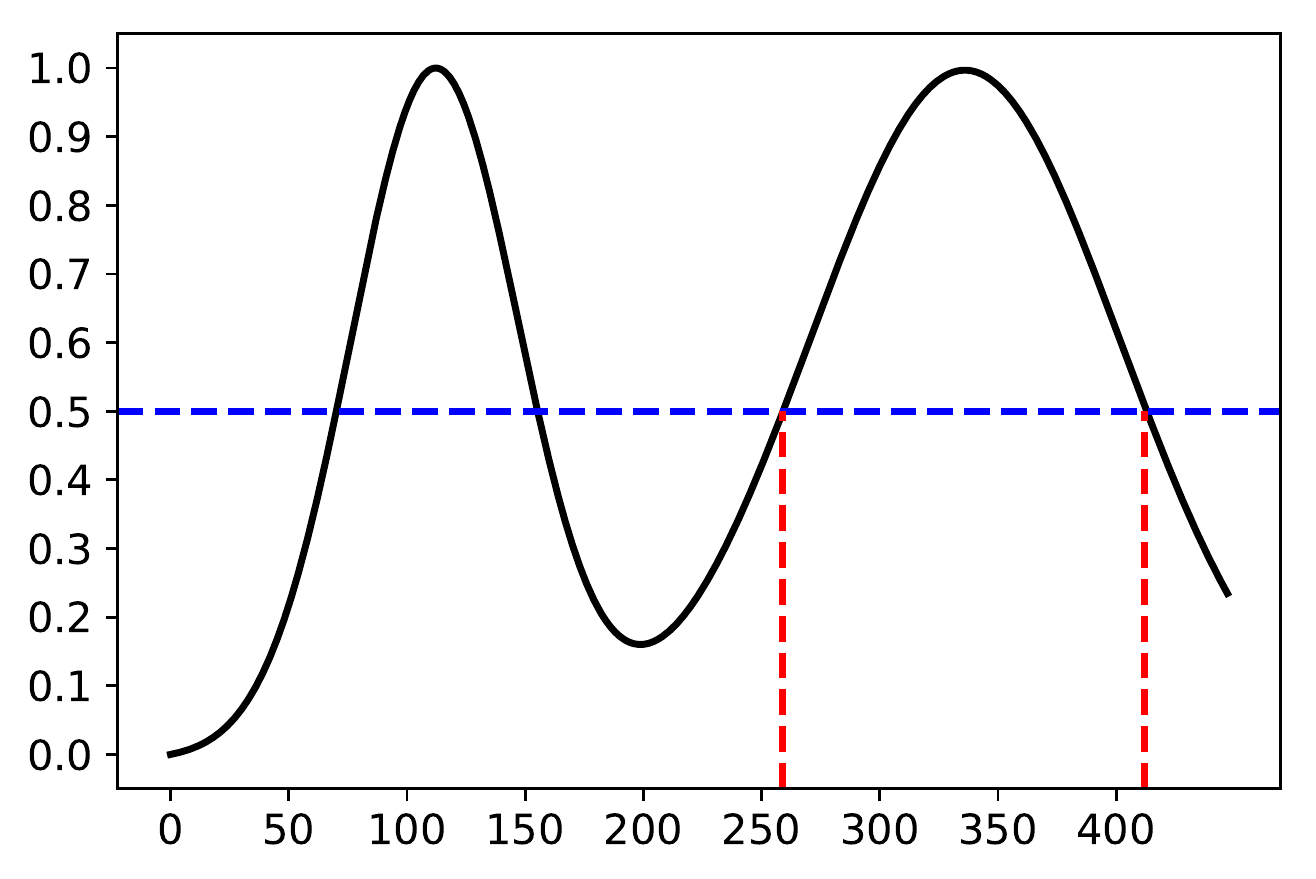}\label{fig:loceg3}}
\caption{Some examples of margin distribution. Black curves represent the margin distribution, and blue dash is the threshold $\tau$, and the best interval between two red dashes is the desirable localization. } \label{fig:loceg}
\end{figure}

The value of $F^i(x,y)$ represents a probability that it belongs to the $i$-th category at spatial location $(x,y)$. 
In order to efficiently localize regions of interest, we decompose each selective attentional map $F^i$ into a row and a column marginal distribution, which represents a probability distribution of objects present at the corresponding location (as shown in Fig.~\ref{fig:mcar}). We compute the marginal distribution based on the class attentional map $F^i$ over $x$ and $y$ axis, respectively, which is
 \begin{equation}\label{eq:margin}
 \begin{aligned}
 \vec p'_x &= \max_{1\leq y \leq h}F^i(x,y), \\
 \vec p'_y &= \max_{1\leq x \leq w}F^i(x,y).
 \end{aligned}
 \end{equation}
 Then, $\vec p'_x$ and $\vec p'_y$ are normalized by min-max normalization such that the distribution is scaled to the range in $[0,1]$ , that is
 \begin{equation}
 \begin{aligned}
\vec p_x &= \big(\vec p'_x - \min_i ({p'_x}^i)\big)/\big (\max_i ({p'_x}^i)- \min_i ({p'_x}^i)\big), \\
\vec p_y &= \big(\vec p'_y - \min_j ({p'_y}^j)\big)/\big(\max_j ({p'_y}^j)- \min_j ({p'_y}^j) \big),
 \end{aligned}
 \end{equation}
where ${p'_x}^i$ represents the $i$-th element of $\vec p'_x$. In order to localize one discriminative region, we need to solve the following integer inequalities:
 \begin{equation}\label{eq:pxpytau}
 \begin{aligned}
 &{p_x}^i  \geq \tau, &s.t. ~~i&=\{1,2,\cdots,w\},\\
 &{p_y}^j  \geq \tau, &s.t. ~~j&=\{1,2,\cdots, h\},
  \end{aligned}
 \end{equation}
where $\tau \in (0,1)$ is a constant threshold. The solution of Eq.~\ref{eq:pxpytau} may be a single interval or a union of multiple ones, and each interval corresponds to the spatial location of a specific object region. The fact is that $\vec p_x$ or $\vec p_y$ may have one peak when input image only contains an object in Fig.~\ref{fig:loceg1} and also may have multiple peaks when input image consists of multiple objects of the same category at different spatial locations in Fig.~\ref{fig:loceg2} and~\ref{fig:loceg3}. However, our objective is to recognize multi-class objects in a given image, and only one discriminative region needs to be selected for each category. Therefore, some constraints have to be added such that a unique interval among multiple feasible intervals can be chosen. To achieve this goal, we pick the interval contained in the global maximum peak for the case of multiple local maximum peaks as shown in Fig.~\ref{fig:loceg2} and choose the widest interval for multiple global maximum peaks as shown in Fig.~\ref{fig:loceg3}. For all selected $topN$ class attentional maps, $topN$ discriminative regions would be generated by solving the Eq.~\ref{eq:pxpytau} conditioned on the above constraints.

\subsection{Two-Stream Learning}
Given a training dataset $\{I_i, \vec {y_i}\}_{i=1}^M$, which $I_i$ is the $i$-th image and $\vec {y_i}=[y_i^1, \cdots, y_i^C]^T$ represents the corresponding labels. The learning goal of our framework
is to find $\vec \theta$, $W$ and $\vec b$ via jointly learning global and local streams in an end-to-end manner. Thus, our overall loss function is formulated as the weighted sum of two streams,
 \begin{equation}\label{eq:loss}
 \mathcal L = \mathcal {L}_g + \mathcal {L}_l,
 \end{equation}
 where $\mathcal {L}_g$ and $\mathcal{L}_l$ represent the global and the local loss, respectively. Specifically, we adopt the binary cross entropy loss for global and local stream,
  \begin{equation}
   \begin{aligned}
\mathcal {L}_g &= \sum_{i=1}^{M}\sum_{j=1}^{C} y_i^j \log(\hat {y_g}_i^j) + (1- y_i^j)\log(1-\hat {y_g}_i^j)\\
\mathcal {L}_l &= \sum_{i=1}^{M}\sum_{j=1}^{C} y_i^j \log(\hat {y_l}_i^j) + (1- y_i^j)\log(1-\hat {y_l}_i^j),
 \end{aligned}
 \end{equation}
 where $\hat {y_g}_i^j$ and $\hat {y_l}_i^j$ are the prediction scores of the $j$-th category of the $i$-th image from global and local streams, respectively. Optimization is performed using SGD and standard back propagation.

\section{Experiments}\label{exps}
In this section, we firstly report extensive experimental results and comparisons that demonstrate the effectiveness of the proposed method. Then, we present ablation studies to carefully evaluate and discuss the contribution of the crucial components in our MCAR. 

\begin{table*}[t]
\centering
\caption{Comparisons of mAP, CP, CR, CF1 and OP, OR, OF1 in $\%$ of our model and state-of-the-art methods on the MS-COCO dataset. * indicates that the results are reproduced by using the open-source code~\cite{chenlearning}, and - denotes the corresponding result is not provided.}\label{table:coco2014}
\footnotesize
\resizebox{\textwidth}{!}{ 
\begin{tabular}{|c|c|c||c||c|c|c|c|c|c||c|c|c|c|c|c|}
\hline
\multirow{2}{*}{Methods} &\multirow{2}{*}{Input Size}  &\multirow{2}{*}{Backbone}  &\multirow{2}{*}{mAP} & \multicolumn{6}{c||}{{All}} &\multicolumn{6}{c|}{{Top3}} \\
\cline{5-16}  & & & &CP &CR &CF1 &OP &OR &OF1 &CP &CR &CF1 &OP &OR &OF1\\
\hline
CNN-RNN~\cite{wang2016cnn}  &-           &VGG16 &61.2 &– &– &– &– &– &–         &66.0 &55.6 &60.4 &69.2 &66.4 &67.8\\
RDAL~\cite{wang2017multi}        &  -         &VGG16 &–      &– &– &– &– &– &–         &79.1 &58.7 &67.4 &84.0 &63.0 &72.0\\
Order-Free RNN~\cite{chen2018order} &- &ResNet-152 &–     &– &– &– &– &– &–    &71.6 &54.8 &62.1 &74.2 &62.2 &67.7\\
ML-ZSL~\cite{lee2018multi}           &-         &ResNet-152 &–    &– &– &– &– &– &–     &74.1 &64.5 &69.0 &–& – &–\\
SRN~\cite{zhu2017learning}        &224$\times$224          &ResNet-101  &77.1 &81.6 &65.4 &71.2 &82.7 &69.9 &75.8 &85.2 &58.8 &67.4 &87.4 &62.5 &72.9\\
ACfs~\cite{guo2019visual}            &288$\times$288        &ResNet-101  &77.5 &77.4 &68.3 &72.2   &79.8 &73.1 &76.3  &85.2 &59.4 &68.0 &86.6 &63.3 &73.1 \\
PLA~\cite{yazici2020orderless}            &288$\times$288        &ResNet-101  &– &80.4 &68.9 &74.2   &81.5 &73.3 &77.1  &–&–&–&–&– &–\\
ResNet-101~\cite{ge2018multi}       &448$\times$448       &ResNet-101  &–     &73.8 &72.9 &72.8 &77.5 &75.1 &76.3 &78.3 &63.7 &69.5 &83.8 &64.9  &73.1\\
Multi-Evidence~\cite{ge2018multi}   &448$\times$448      &ResNet-101  &– &80.4 &70.2 &74.9 &85.2 &72.5 &78.4 &84.5 &62.2 &70.6 &89.1 &64.3 &74.7\\
SSGRL*~\cite{chenlearning}            &448$\times$448        &ResNet-101 &81.9 &84.2 &70.3 &76.6 &85.8 &72.4 &78.6 &88.0 &63.1 &73.5 &90.2 &64.5 &75.2\\
\hline\hline
MCAR &288$\times$288 &ResNet-101    &80.5 &81.8 &69.2 &75.0 &84.9 &72.2 &78.0  &85.8 &62.6 &72.4 &88.9 &64.7 &74.9 \\
\hline\hline
Baseline &448$\times$448&ResNet-101  &77.1 &72.7 &72.3 &72.5 &77.4 &75.5 &76.5  &77.8 &63.5 &69.9 &84.0 &65.5 &73.6\\
MCAR &448$\times$448&ResNet-101  &\textbf{83.8} &\textbf{85.0} &\textbf{72.1} &\textbf{78.0} &\textbf{88.0} &\textbf{73.9} &\textbf{80.3} &\textbf{88.1} &\textbf{65.5} &\textbf{75.1} &\textbf{91.0} &\textbf{66.3} &\textbf{76.7}\\
\hline\hline
SSGRL~\cite{chenlearning}        & 576$\times$576           &ResNet-101  &83.8 &\textbf{89.9} &68.5 &76.8 &\textbf{91.3} &70.8 &79.7 &\textbf{91.9} &62.5 &72.7 &\textbf{93.8} &64.1 &76.2 \\
MCAR &576$\times$576 &ResNet-101 &\textbf{84.5}  &84.3 &\textbf{73.9} &\textbf{78.7} &86.9 &\textbf{76.1} &\textbf{81.1} &87.8 &\textbf{65.9} &\textbf{75.3} &90.4 &\textbf{67.1} &\textbf{77.0}\\
\hline
\end{tabular}}
\end{table*}

\subsection{Experiment Setting}
\noindent \textbf{Implementation Details.}
We perform experiments to validate the effectiveness of the proposed MCAR on three benchmarks in multi-label classification: MS-COCO~\cite{lin2014microsoft}, PASCAL VOC 2007 and 2012~\cite{everingham2010pascal}, using the open-source framework PyTorch. 

Following recent MLR works, we compare the proposed method with state-of-the-arts using the powerful ResNet-50 and ResNet-101~\cite{he2016deep} models. Some popular and lightweight models, such as MobileNet-v2~\cite{sandlermobilenetv2}, are also used to further evaluate our method. In general, for each of these networks we remove the fully-connected layers before the final output and replace them with global pooling followed by a 1$\times$1 convolutional layer and a sigmoid layer. These models are all pre-trained on ImageNet and we train them using image-level labels only. The stochastic gradient descent (SGD) optimizer is used with the momentum of 0.9 and the weight decay of $0.0001$. The initial learning rate is set to 0.001 for all layers but 0.01 for the 1$\times$1 convolution, and they are decreased by a factor of 10 in the $30^{th}$ and $50^{th}$ epoch and the network is trained for 60 epochs in total.

During training, all input images are resized into a fixed size (\ie, 256$\times$256 or 448$\times$448) with random horizontal flips and color jittering for data augmentation. In order to speed up the convergence of the network, we don't use the random crop although it can bring performance improvement but need more training time. Unless otherwise stated, we set $topN$ as 4 and $\tau$ as 0.5 in our experiments. The effects of hyper-parameters ($topN$ and $\tau$) is discussed in Section~\ref{exps:as}.

\noindent \textbf{Evaluation Metrics.} 
The performance of MLR mainly employ two metrics which are 
the average precision (AP) for each category and the mean average precision (mAP) overall categories. We first employ AP and mAP to evaluate all the methods.  Following conventional setting~\cite{wei2015hcp,chen2019multi,chenlearning},  we also compute the precision, recall and F1-measure for comparison performance on MS-COCO dataset. For each image, we assign a positive label if its prediction probability is greater than a threshold~(0.6) and compare them with the ground-truth labels. The overall precision~(OP), recall~(OR), F1-measure (OF1) and per-category precision~(CP), recall~(CR), F1-measure (CF1) are computed as follows:

 \begin{equation}
 \begin{aligned}
\mathrm{OP} &= \frac{\sum_i M_c^i}{\sum_i M_p^i}, &\mathrm{OR} &= \frac{\sum_i M_c^i}{\sum_i M_g^i}, \\
\mathrm{CP}&=\frac{1}{C}\sum_i \frac{M_c^i}{M_p^i}, &\mathrm{CR}&=\frac{1}{C}\sum_i \frac{M_c^i}{M_g^i}, \\
\mathrm{OF1}&=\frac{2*\mathrm{OP}*\mathrm{OR}}{\mathrm{OP}+\mathrm{OR}},
&\mathrm{CF1}&=\frac{2*\mathrm{CP}*\mathrm{CR}}{\mathrm{CP}+\mathrm{CR}},
 \end{aligned}
 \end{equation} 
where $M_c^i$ is the number of images correctly predicted for the $i$-th category,  $M_p^i$ is the number of predicted images for the $i$-th category, $M_g^i$ is the number of ground truth images for the $i$-th category. We also compute these above metrics via another way that each image is assigned labels with top3 highest score. It is worthy to notice that these metrics may be affected by the threshold.  Among these metrics, OF1 and CF1 are more stable than OP, CP, OR and CR. AP and mAP are the most important metrics which can provide a more comprehensive comparison.

\subsection{Comparisons with State-of-the-Arts}\label{cstoa}
To verify the effectiveness of our method, we compare the proposed method with state-of-the-arts on MS-COCO~\cite{lin2014microsoft}  and PASCAL VOC 2007 \& 2012~\cite{everingham2010pascal}.

\noindent \textbf{MS-COCO.}
MS-COCO~\cite{lin2014microsoft} is a widely used dataset to evaluate multiple tasks such as object detection, semantic segmentation and image caption, and it has been adopted to evaluate multi-label image recognition recently. 
It contains 82,081 images as the training set and 40,137 images as validation set and covers 80 object categories.
Compared to VOC 2007~\& 2012~\cite{everingham2010pascal}, both the size of training set and the number of object categories are increased. Meanwhile, the number of labels of different images, the scale of different objects and the number of images in each category vary considerably, which makes it more challenging.

 \begin{figure*}[t]
 \centering
   \vspace{0pt}
 {\includegraphics[width= 0.95\textwidth]{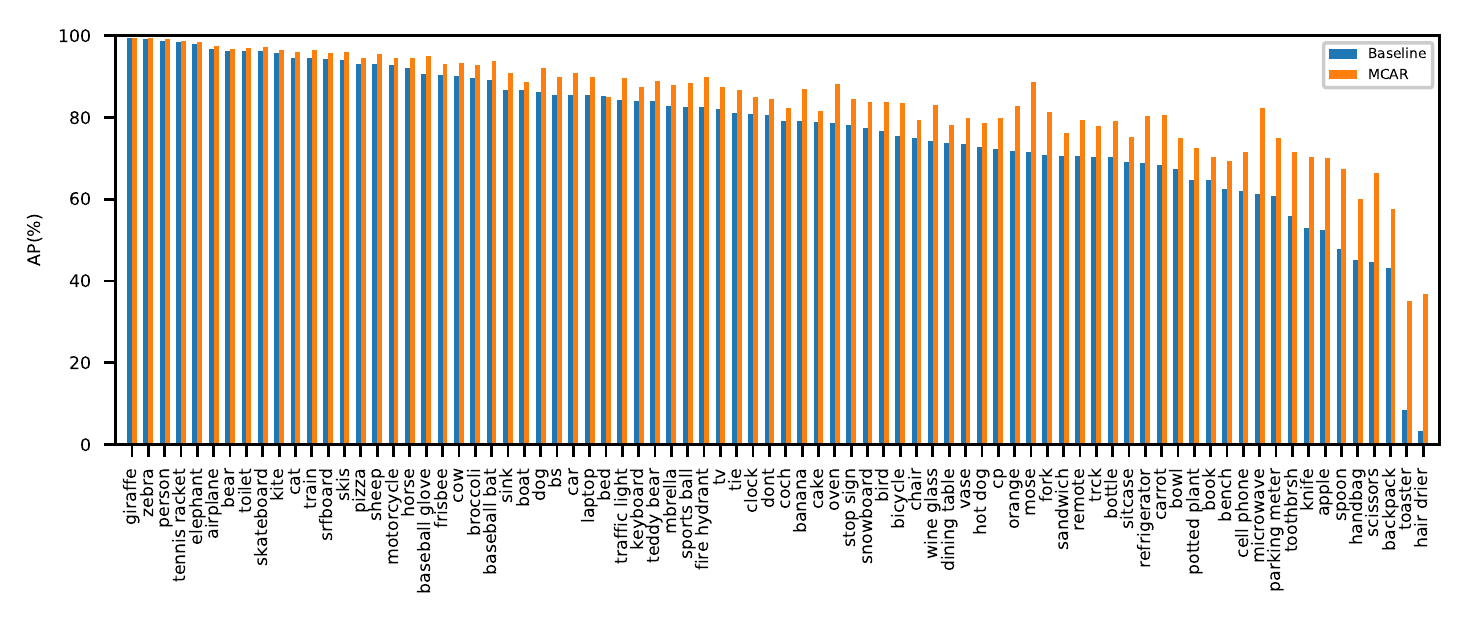}}
  \vspace{0pt}
\caption{AP (in $\%$) of each category of our proposed framework and the ResNet-101 baseline on MS-COCO dataset. Our MCAR has significant improvements on almost all categories, especially for some difficult categories such as ``toaster" and ``hair drier". } \label{fig:cocap}
\end{figure*}

\noindent \textbf{Results on MS-COCO.}
The results on MS-COCO are reported in Table~\ref{table:coco2014}. When the input size is 448$\times$448 (the most common setting in MLR), our method is already comparable to the state-of-the-art SSGRL~\cite{chenlearning} which uses additional label dependency and larger input to boost performance. Moreover, if we simply resize the input image to 576$\times$576 during the testing stage while still using the model weights trained with 448$\times$448 inputs, our method achieves 84.5\% mAP which outperforms the SSGRL by 0.7\%.
In order to fairly compare with the SSGRL, we re-implement the experiment with 448$\times$448 input following the same setting as described in the SSGRL. In Table~\ref{table:coco2014}, we can see that our method significantly beats the SSGRL and improves it by 1.9 points (83.8\% vs. 81.9\%). Note that PLA~\cite{yazici2020orderless} models label correlation through exploiting LSTM model. Using the same input size~(288$\times$288), our method gets higher F1 scores than PLA, which further indicates that it is very important to exploit image semantics for multi-label image recognition.

The performance of our method is also significantly better than that of Multi-Evidence~\cite{ge2018multi}, and it improves CF1 by 3.1\%, OF1 by 1.9\%, CF1-top3 by 4.5\%, OF1-top3 by 2.0\%.  
Note that our baseline ResNet-101 model achieves 77.1\% mAP, which should be close to that of the baseline of Multi-Evidence~\cite{ge2018multi} because of nearly the same F1-measures. 
In comparison to the baseline, our method is 6.7\% higher in mAP~(83.8\% vs. 77.1\%).

Meanwhile, we show the AP performance of each class for further comparison with the baseline model in Fig~\ref{fig:cocap}. It is obvious that our method has significant improvements on almost all categories, especially for some difficult categories such as ``toaster" and ``hair drier". In short, MCAR outperforms all state-of-the-art methods and significantly surpasses the baseline by a large margin even though it does not need a large number of proposals or label dependency information. This further demonstrates the effectiveness of the proposed method for large-scale multi-label image recognition.

 \begin{table*}[t]
	\centering
	\caption{Comparisons of AP and mAP in $\%$ of our model and state-of-the-art methods on the PASCAL VOC 2007. $*$ indicates methods using larger input size (576$\times$576).}\label{table:voc07}
	\footnotesize
	\resizebox{\textwidth}{!}{
		\begin{tabular}{|@{\,}c@{\,}||@{\,}c@{\,}|| *{20}{@{\,}c@{\,}}||@{\,}c@{\,}|}
			\hline
			Methods  &Backbone &aero &bike &bird &boat &bottle &bus &car &cat &chair &cow &table &dog &horse &mbike &person &plant &sheep &sofa &train &tv &mAP\\
			\hline\hline
			CNN-RNN~\cite{wang2016cnn} &VGG16                &96.7 &83.1 &94.2 &92.8 &61.2 &82.1 &89.1 &94.2 &64.2 &83.6 &70.0 &92.4  &91.7 &84.2 &93.7 &59.8 &93.2 &75.3 &\textbf{99.7} &78.6 &84.0\\
			VGG+SVM~\cite{Simonyan15} &~VGG16\&19~  &98.9 &95.0 &96.8 &95.4 &69.7 &90.4 &93.5 &96.0 &74.2 &86.6 &87.8 &96.0 &96.3 &93.1 &97.2 &70.0 &92.1 &80.3 &98.1 &87.0 &89.7\\
			Fev+Lv~\cite{yang2016exploit} &VGG16                  &97.9 &97.0 &96.6 &94.6 &73.6 &93.9 &96.5 &95.5 &73.7 &90.3 &82.8 &95.4 &97.7 &95.9 &98.6 &77.6 &88.7 &78.0 &98.3 &89.0 &90.6\\
			HCP~\cite{wei2015hcp} &VGG16                             &98.6 &97.1 &98.0 &95.6 &75.3 &94.7 &95.8 &97.3 &73.1 &90.2 &80.0 &97.3 &96.1 &94.9 &96.3 &78.3 &94.7 &76.2 &97.9 &91.5 &90.9\\
	                RDAL~\cite{wang2017multi}  &VGG16        &98.6 &97.4 &96.3 &96.2 &75.2 &92.4 &96.5 &97.1 &76.5 &92.0 &87.7 &96.8 &97.5 &93.8 &98.5 &81.6 &93.7 &82.8 &98.6 &89.3 &91.9\\
	                RARL~\cite{chen2018recurrent} &VGG16  &98.6 &97.1 &97.1 &95.5 &75.6 &92.8 &96.8 &97.3 &78.3 &92.2 &87.6 &96.9 &96.5 &93.6 &98.5 &81.6 &93.1 &83.2 &98.5 &89.3 &92.0\\
	                SSGRL*~\cite{chenlearning} &ResNet-101                 &99.5 &97.1 &97.6 &97.8 &82.6 &94.8 &96.7 &98.1 &78.0 &\textbf{97.0} &85.6 &97.8 &98.3 &96.4 &98.1 &\textbf{84.9} &96.5 &79.8 &98.4 &92.8 &93.4\\
	                \hline                                                                         
	                Baseline &ResNet-101                                              &99.0&97.9&97.2&97.6&80.2&93.6&96.0&98.0&81.8&92.0&84.6&97.5&97.2&95.3&97.9&81.8&94.6&84.1&98.2&93.6 &92.9\\

	                MCAR &ResNet-101                                                 &	\textbf{99.7} &\textbf{99.0}& 98.5&\textbf{98.2}&\textbf{85.4}&\textbf{96.9}&\textbf{97.4}&\textbf{98.9}&\textbf{83.7}& 95.5&\textbf{88.8}&\textbf{99.1}& 98.2& 95.1&\textbf{99.1}& 84.8&\textbf{97.1}&\textbf{87.8}& 98.3&\textbf{94.8} &\textbf{94.8} \\ 	    
            \hline
	\end{tabular}}
\end{table*} 
\begin{table*}[t]
	\centering
		\caption{Comparisons of AP and mAP in $\%$ of our model and state-of-the-art methods on the PASCAL VOC 2012. $*$ indicates methods using larger input size (576$\times$576).}\label{table:voc12}
		\footnotesize
		\resizebox{\textwidth}{!}{
			\begin{tabular}{|@{\,}c@{\,}||@{\,}c@{\,}|| *{20}{@{\,}c@{\,}}||@{\,}c@{\,}|}
				\hline
				Methods &Backbone &aero &bike &bird &boat &bottle &bus &car &cat &chair &cow &table &dog &horse &mbike &person &plant &sheep &sofa &train &tv &mAP\\
				\hline
				VGG+SVM~\cite{Simonyan15} &VGG16\&19                  &99.0 &89.1 &96.0 &94.1 &74.1 &92.2 &85.3 &97.9 &79.9 &92.0 &83.7 &97.5 &96.5 &94.7 &97.1 &63.7 &93.6 &75.2 &97.4 &87.8 &89.3\\			        
			        	Fev+Lv~\cite{yang2016exploit} &VGG16                         &98.4 &92.8 &93.4 &90.7 &74.9 &93.2 &90.2 &96.1 &78.2 &89.8 &80.6 &95.7 &96.1 &95.3 &97.5 &73.1 &91.2 &75.4 &97.0 &88.2 &89.4\\
				HCP~\cite{wei2015hcp}                                                    &VGG16 &99.1 &92.8 &97.4 &94.4 &79.9 &93.6 &89.8 &98.2 &78.2 &94.9 &79.8 &97.8 &97.0 &93.8 &96.4 &74.3 &94.7 &71.9 &96.7 &88.6 &90.5\\
				SSGRL*~\cite{chenlearning} &ResNet-101                       &99.5 &95.1 &97.4 &96.4 &85.8 &94.5 &93.7 &\textbf{98.9} &86.7 &96.3 &84.6 &\textbf{98.9} &\textbf{98.6} &96.2 &98.7 &82.2 &\textbf{98.2} &\textbf{84.2}&98.1 &93.5 &93.9\\
				\hline
				MCAR &MobileNet-v2                                                     &98.6&92.3&95.4&93.3&77.7&93.8&92.6&97.6&80.8&90.9&82.3&96.5&96.6&95.5&98.3&78.4&92.6&78.7&96.8&90.9 &91.0\\
				MCAR &ResNet-50                                                         &99.6&95.6&97.5&95.2&85.1&95.5&94.3&98.6&85.2&95.8&83.9&98.4&98.0&97.2&98.8&81.6&95.5&81.8&98.3&\textbf{93.6}&93.5\\
				MCAR &ResNet-101                                                       &\textbf{99.6} &\textbf{97.1}&\textbf{98.3}&\textbf{96.6}&\textbf{87.0}&\textbf{95.5}&\textbf{94.4}&98.8&\textbf{87.0}&\textbf{96.9}&\textbf{85.0}&98.7&98.3&\textbf{97.3}&\textbf{99.0}&\textbf{83.8}&96.8&83.7&\textbf{98.3}&93.5  &\textbf{94.3}\\
				\hline
	\end{tabular}}
\end{table*}
\noindent \textbf{PASCAL VOC 2007 and 2012.}
PASCAL VOC 2007 and 2012~\cite{everingham2010pascal} are the most widely used datasets for MLR. There are 9,963 and 22,531 images in  VOC 2007 and 2012, respectively. Each image contains one or several labels, corresponding to 20 object categories. These images are divided into three parts including~\emph{train}, \emph{val} and~\emph{test} sets. In order to fairly compare with other competitors, we follow the common setting to train our model on the~
\emph{train-val} sets, and then evaluate produced models on the~\emph{test} set. VOC 2007 contains a~\emph{train-val} set of 5,011 images and a~\emph{test} set of 4,952 images. VOC 2012 consists of 11,540 images as~\emph{train-val} set and 10,991 images as the~\emph{test} set.

\noindent \textbf{Results on VOC 2007.}
We first report the AP for each category and the mAP for all categories on VOC 2007~\emph{test} set in Table~\ref{table:voc07}. The current state-of-the-art is SSGRL~\cite{chenlearning} which uses GCN to model label dependency to boost the performance. We can see that our method achieves the best mAP performance among all methods. 
It largely outperforms the SSGRL~\cite{chen2019multi} by 1.4 points~(94.8\% vs. 93.4\%) when SSGRL uses a larger input size 576$\times$576. 
Moreover, the proposed method improves the baseline ResNet-101 model by 1.9\% under the same setting such as data augmentation and hyper-parameters of optimization. Last but not least, our framework shows good performance for some difficult categories such as ``bottle", ``table" and ``sofa". This shows that exploiting global and local vision information is very crucial for multi-label recognition.

\noindent \textbf{Results on VOC 2012.}
We report the results on VOC 2012~\emph{test} set with PASCAL VOC evaluation server in Table~\ref{table:voc12}. We compare state-of-the-arts with our method on several backbone networks. First, we still win the best mAP performance with a smaller input size compared to SSGRL~\cite{chenlearning} when ResNet-101 is considered as a backbone. Second, our method achieves better performance using lightweight networks, \ie MobileNet-v2 and ResNet-50, than that of VGG. This implies that it may be easy to extend our method to resource-restricted devices such as mobile phones.

\subsection{Ablation Study}\label{exps:as}
In order to comprehend how MCAR works, we perform exhaustive experiments to analyze the components in MCAR.
We firstly analyze the contribution of each component in our two-stream architecture and demonstrate its effectiveness. The training details are exactly the same as those described in Section~\ref{cstoa}. Then, the effect of the attentional maps selection criteria and learning strategy is analyzed. Next, we also present the effects of MCAR in different hyper-parameters ($topN$ and $\tau$) appearing in the local region localization module. The experiment is conducted on VOC 2007 and MS-COCO using different backbone networks, \eg MobileNet-v2, ResNet-50 and ResNet-101, and we set the input size to 256$\times$256. 
Finally, we extensively analyze the effects of our method under different conditions such as different global pooling strategies, various input sizes, and different network architectures.

\begin{table}[t]
	\centering
	\caption{Ablative study of two streams in MCAR with ResNet-101 backbone and the input size of 448$\times$448.}\label{table:voc-coco}
	\footnotesize{
	\begin{tabular}{|c||c|cc||c||c|}
	\hline
	Line No. &Methods &{Global} &{Local}   &{VOC 2007} &{MS-COCO}\\
	\hline
	{\textcolor[rgb]{0.6,0.6,0.6}{~~~0~~~}}  &Baseline &$\surd$ & &92.9 &77.1\\
	\hline
	{\textcolor[rgb]{0.6,0.6,0.6}{~~~1~~~}}   &\multirow{3}*{MCAR}&$\surd$ & &93.4  {\color{red} $\uparrow$0.5}   &81.9  {\color{red} $\uparrow$4.8} \\
	{\textcolor[rgb]{0.6,0.6,0.6}{~~~2~~~}}   & & &$\surd$ &94.2 {\color{red} $\uparrow$1.3}  &82.9 {\color{red} $\uparrow$5.8}  \\
         {\textcolor[rgb]{0.6,0.6,0.6}{~~~3~~~}}   & &$\surd$ &$\surd$ &94.8 {\color{red} $\uparrow$1.9}  &83.8 {\color{red} $\uparrow$6.7}  \\
        \hline
       \end{tabular}}
\end{table}

\noindent \textbf{Contributions of proposed global-to-local framework.}  
To explore the effectiveness of two streams, we jointly train the global and local streams in MCAR, and during the inference stage, we report the influence of using each stream in Table~\ref{table:voc-coco}. Firstly, thanks to the joint training strategy, our MCAR significantly outperforms the baseline method even when the same global image is taken as input (line 1 vs. line 0). Such improvement is very intuitive because MCAR is more robust and generalized by learning on not only global image but also various scales of local regions. Secondly, we can see that using local stream alone performs better than only using global stream (line 2 vs. line 1), which is because the local stream is able to flexibly focus on the details of each object. Nonetheless, we want to emphasize that the global stream plays an important role in guiding the learning of local stream. Last but not least, it is obvious that employing both global and local streams achieves the best results (line 3). This is similar to humans perception because we usually make a final decision after our brain gathers information from different spatial locations and object scales.

\begin{figure*}[t]
 \centering
 \vspace{-5pt}
 \subfloat[MobileNet-v2]
 {\includegraphics[width= 0.3\columnwidth]{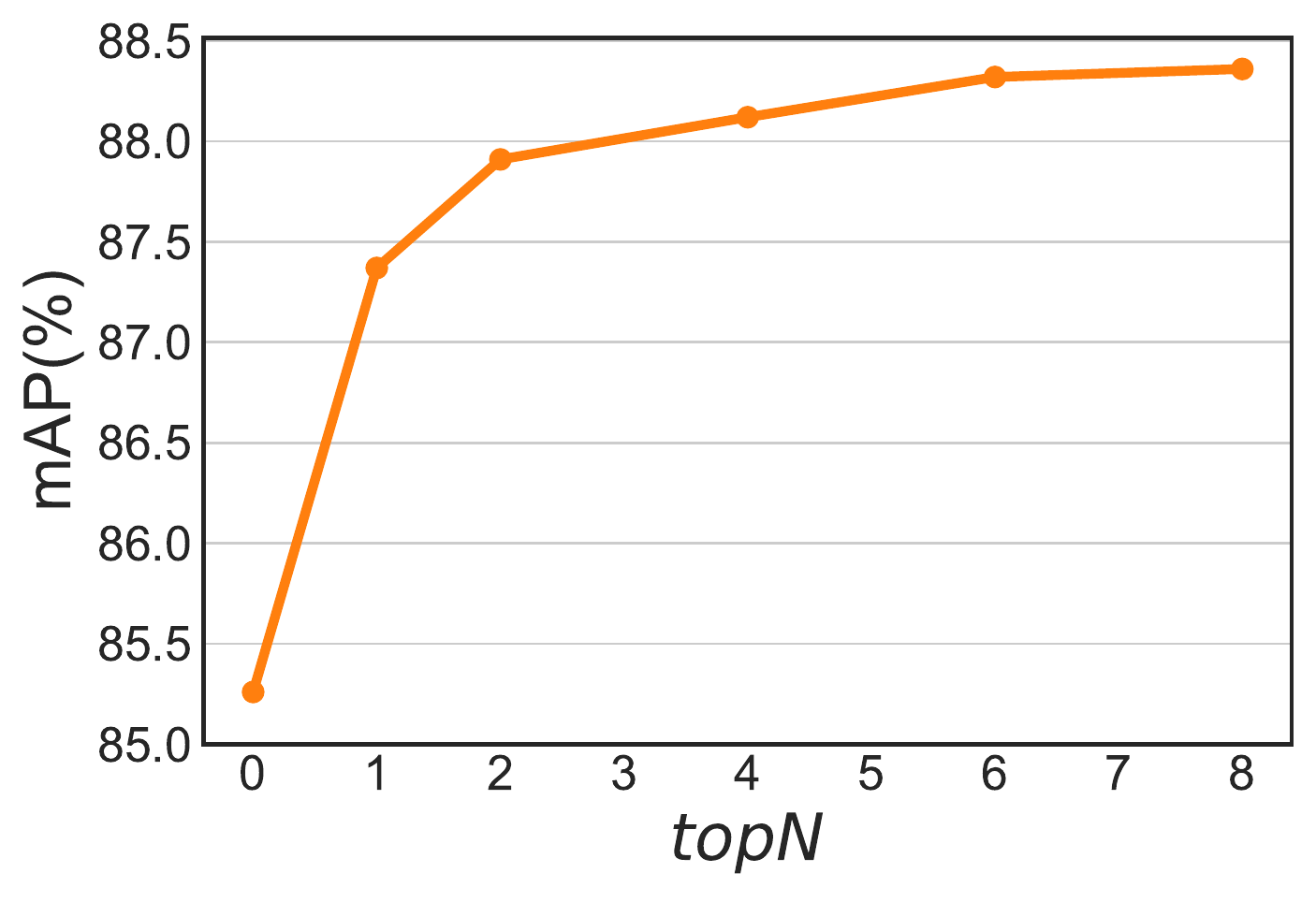}\label{fig:voc-m2topn}}
 \subfloat[ResNet-50]
 {\includegraphics[width= 0.3\columnwidth]{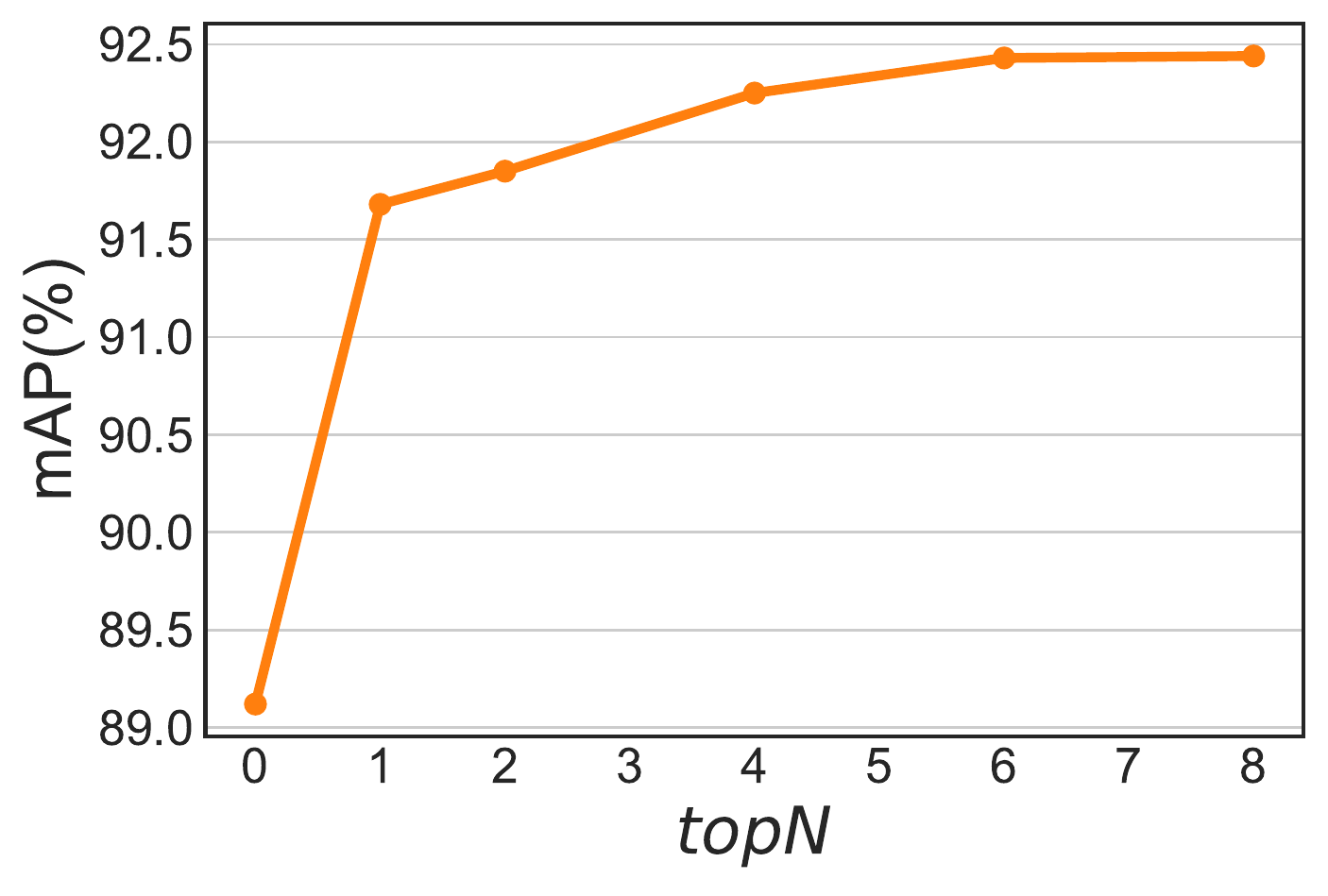}\label{fig:voc-res50topn}}
  \subfloat[ResNet-101]
 {\includegraphics[width= 0.3\columnwidth]{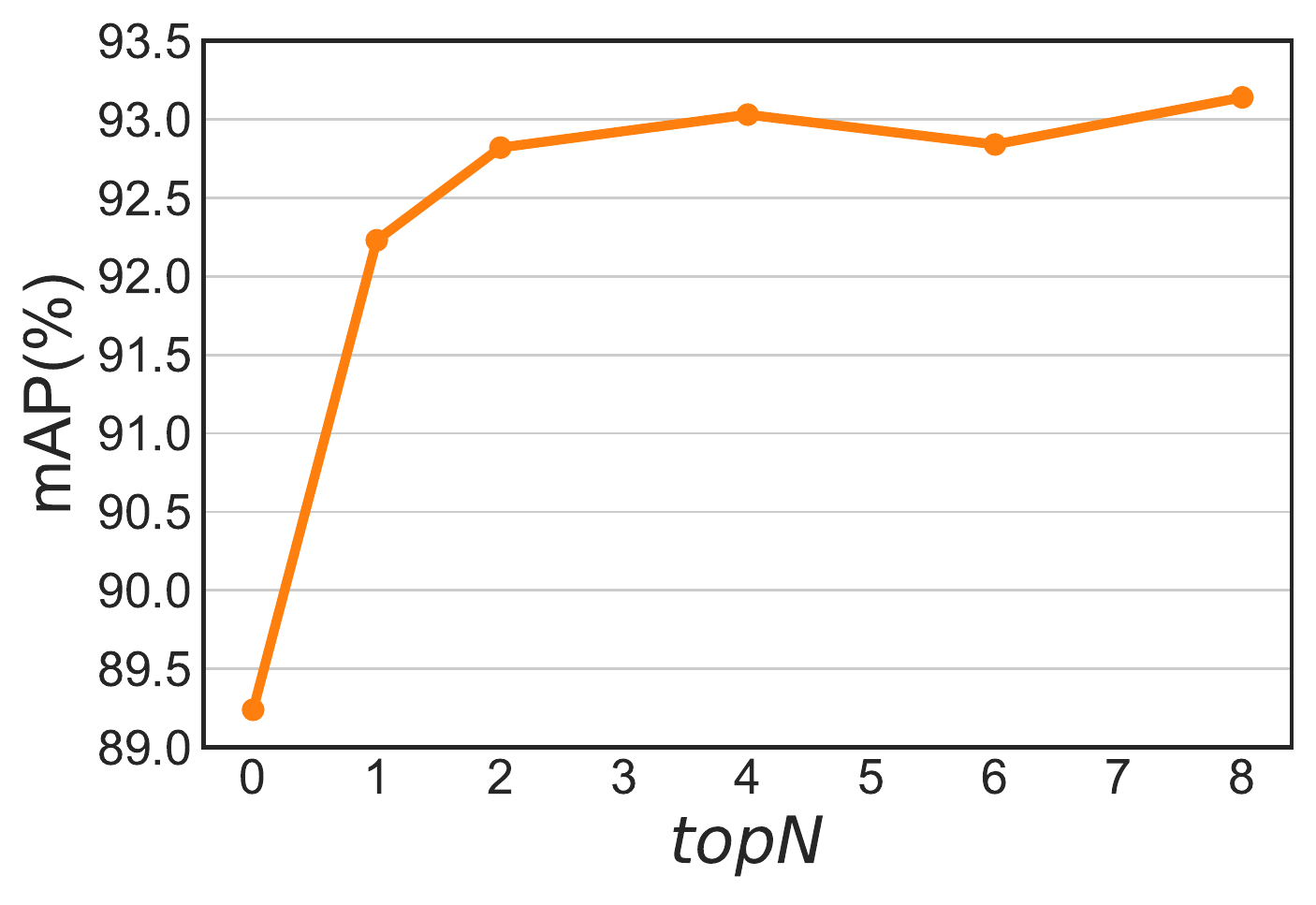}\label{fig:voc-res101topn}}
 \quad \vrule \quad
 \subfloat[MobileNet-v2]
 {\includegraphics[width= 0.3\columnwidth]{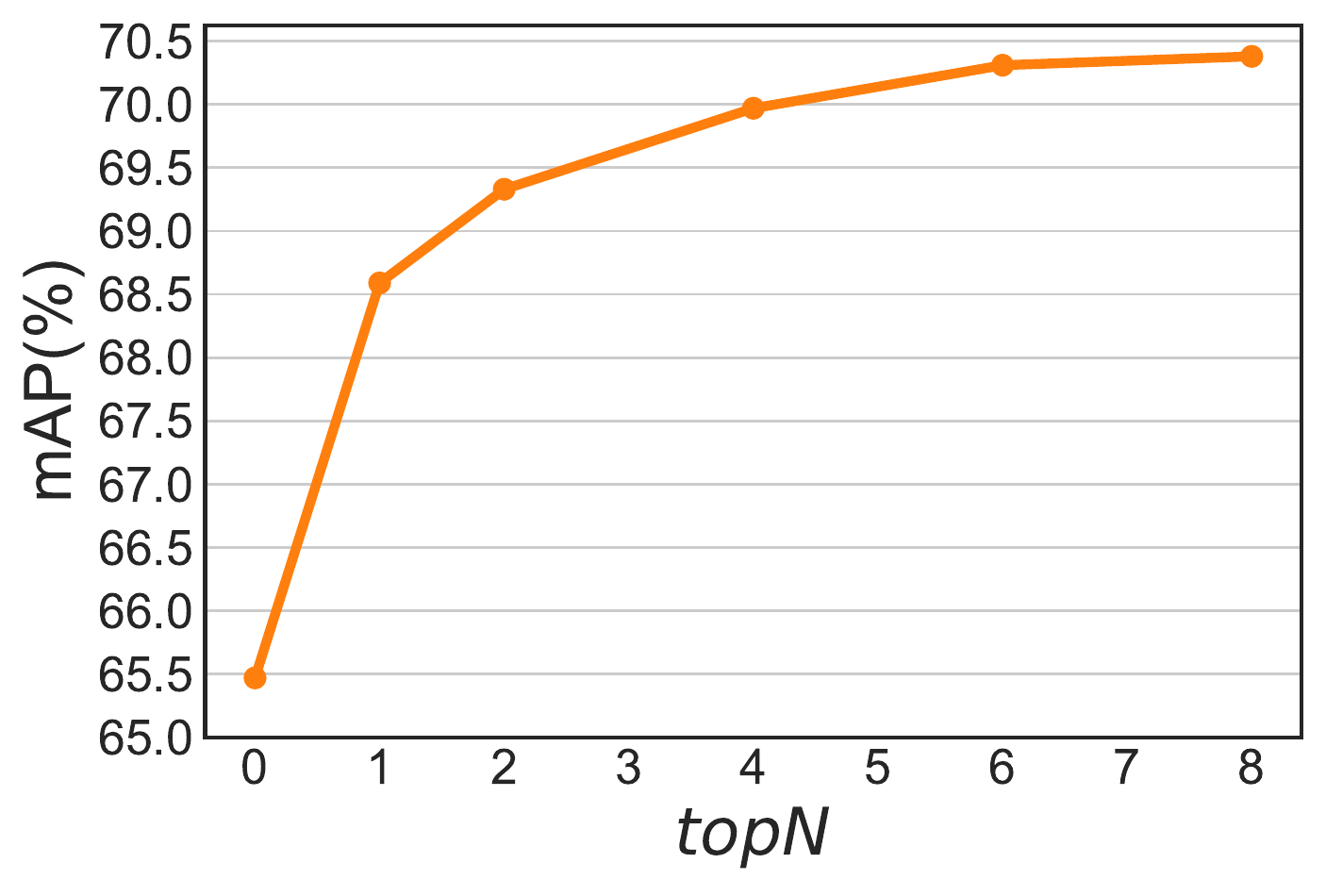}\label{fig:coco-m2topn}}
 \subfloat[ResNet-50]
 {\includegraphics[width= 0.3\columnwidth]{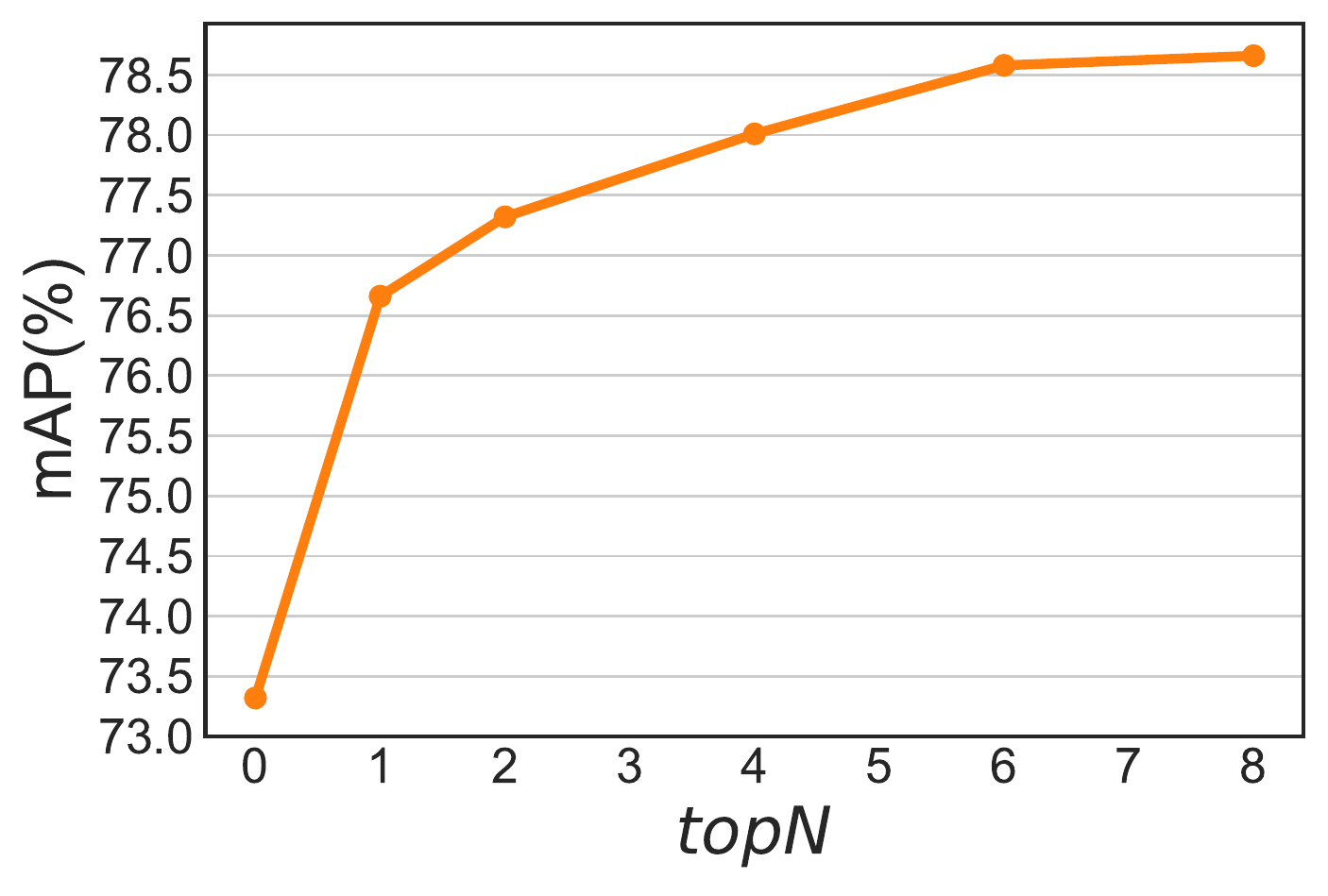}\label{fig:coco-res50topn}}
  \subfloat[ResNet-101]
 {\includegraphics[width= 0.3\columnwidth]{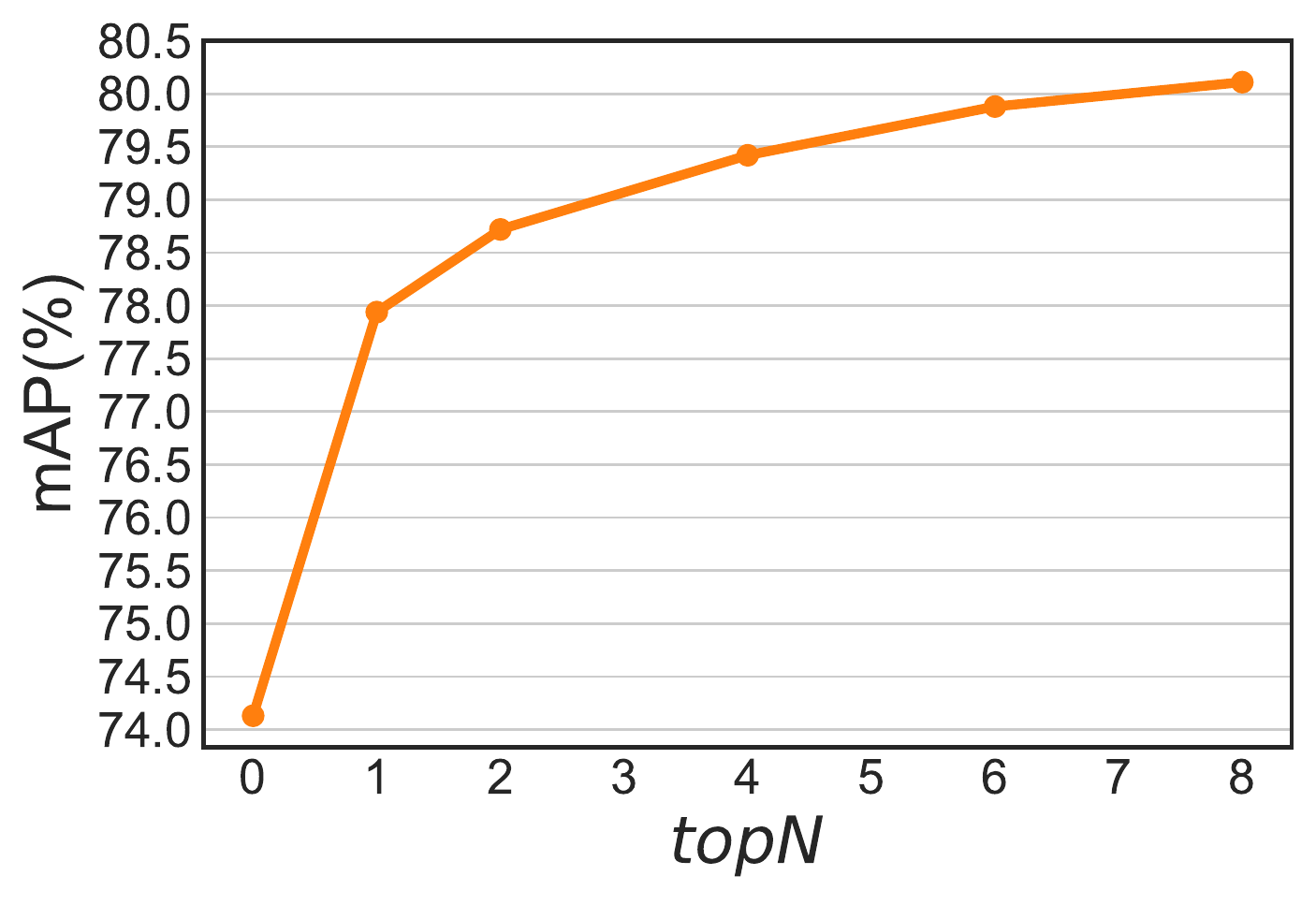}\label{fig:coco-res101topn}}\\
 \subfloat[MobileNet-v2]
 {\includegraphics[width= 0.3\columnwidth]{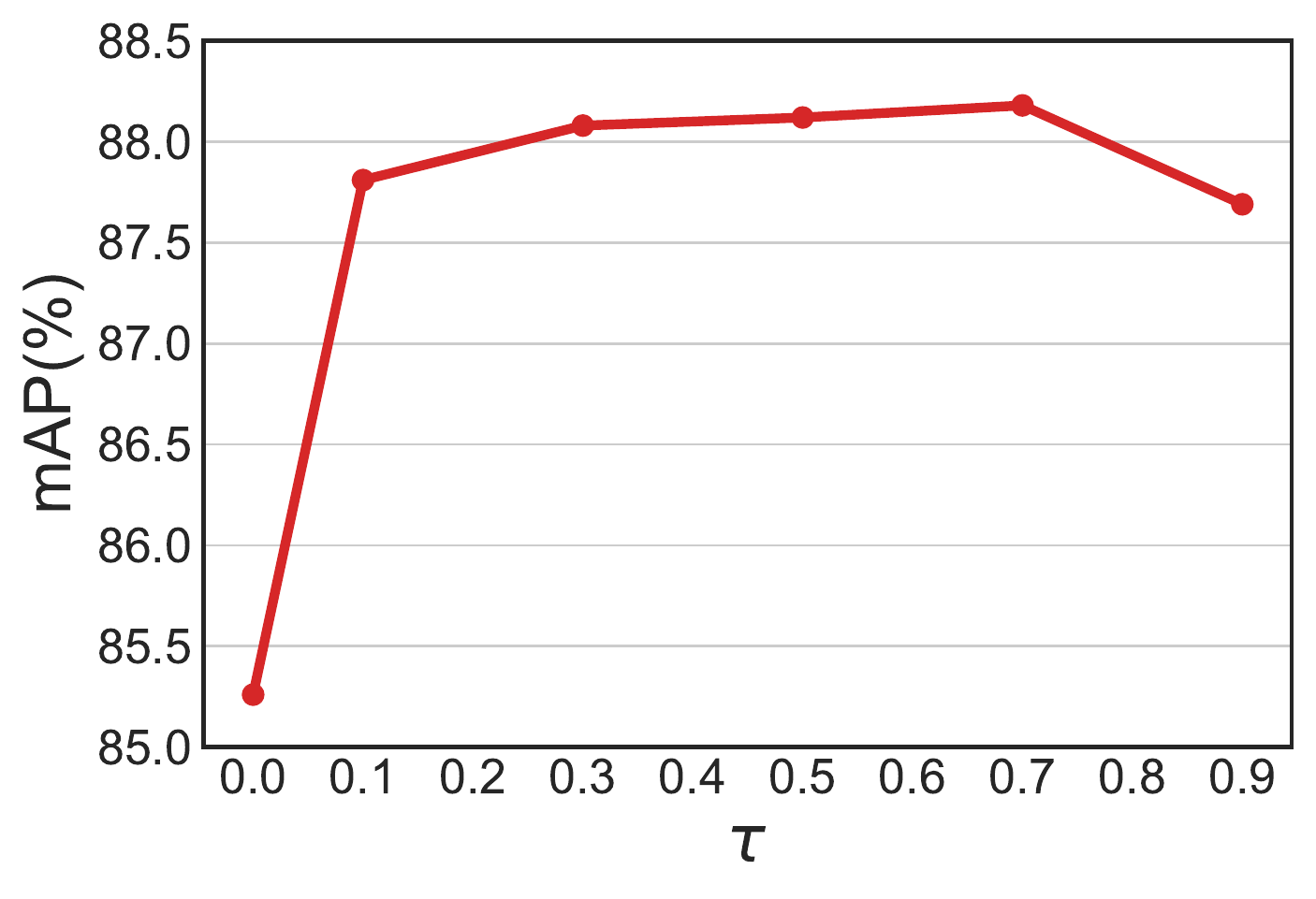}\label{fig:voc-m2thresh}}
 \subfloat[ResNet-50]
 {\includegraphics[width= 0.3\columnwidth]{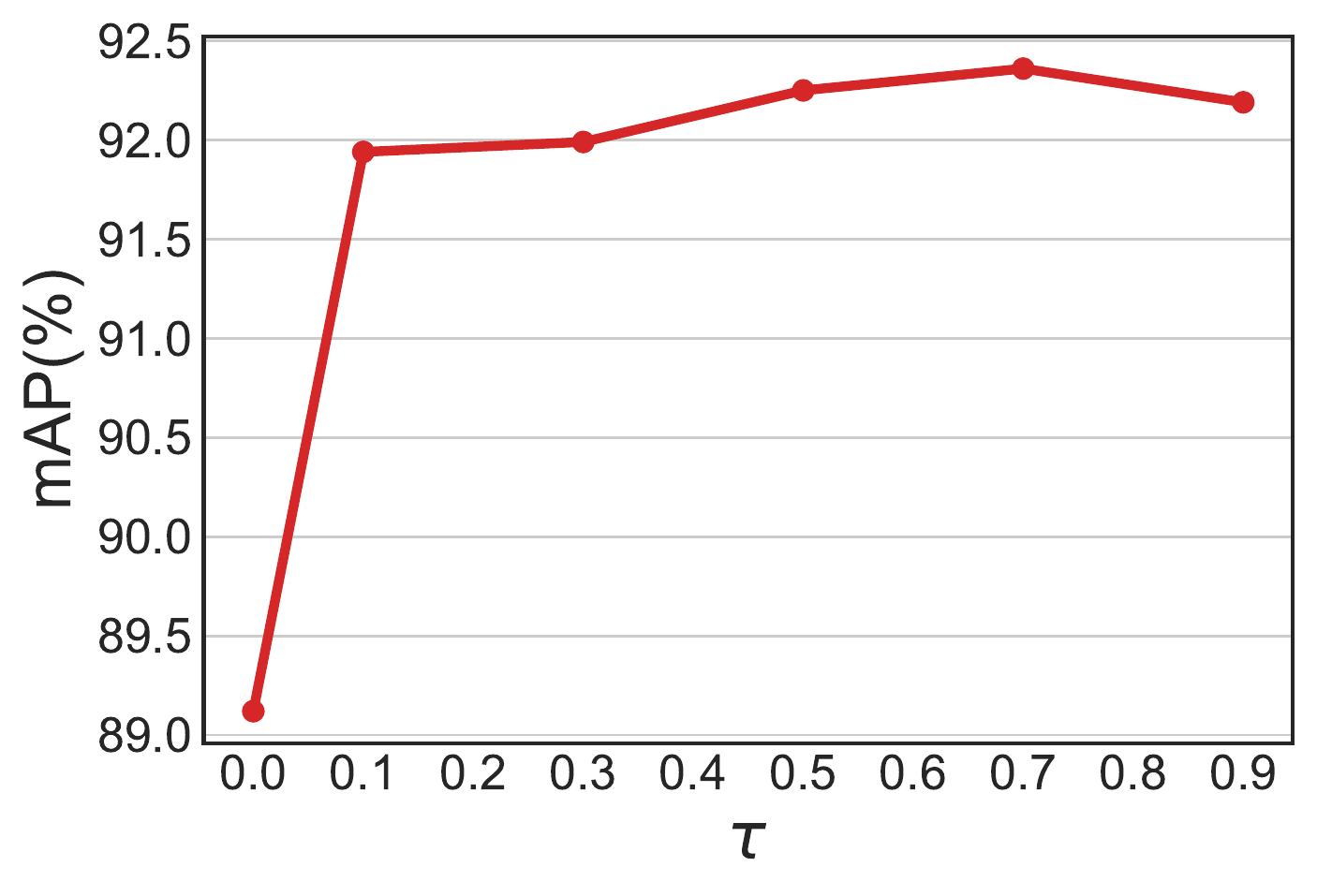}\label{fig:voc-res50thresh}}
 \subfloat[ResNet-101]
 {\includegraphics[width= 0.3\columnwidth]{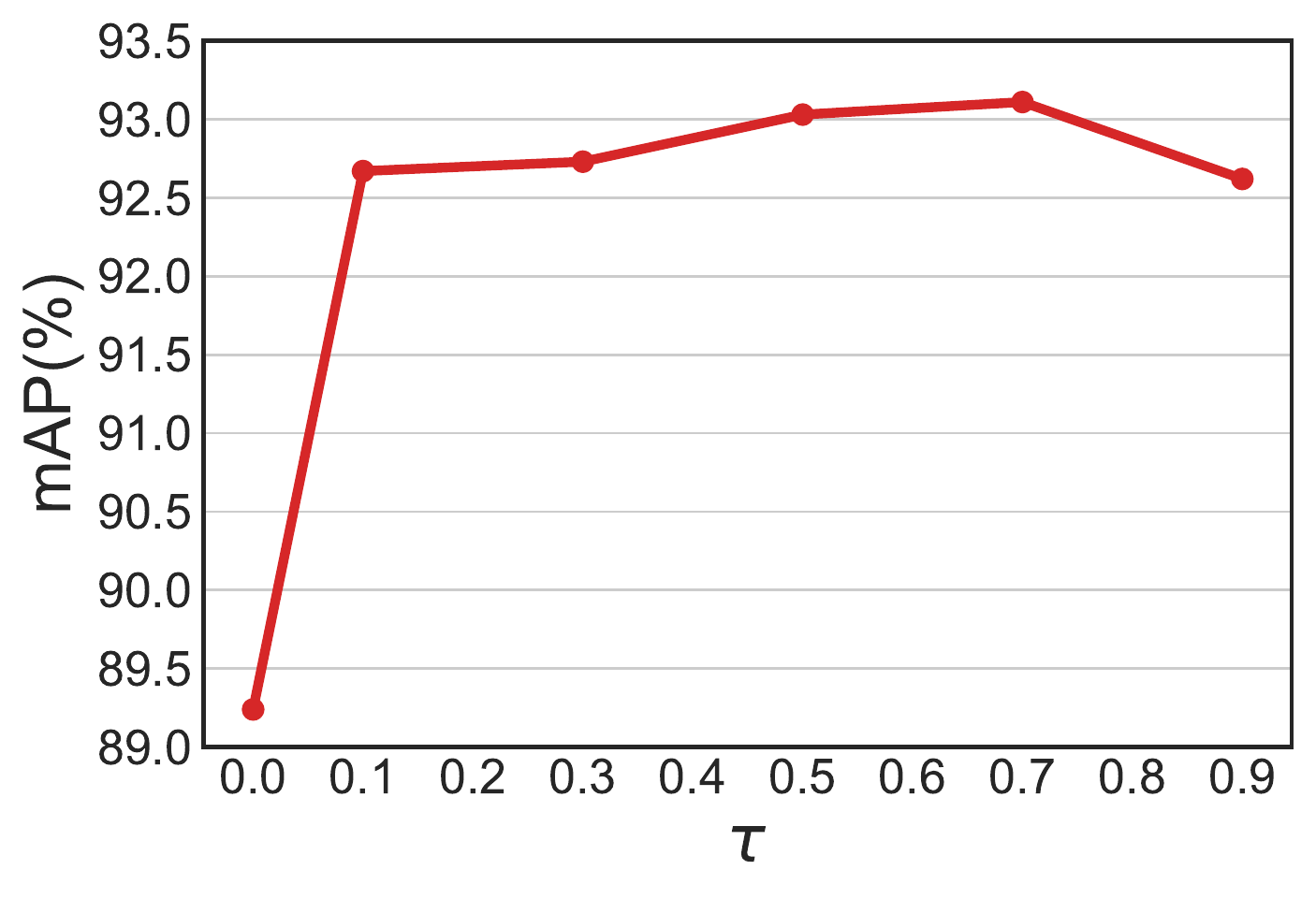}\label{fig:voc-res101thresh}}
  	\quad \vrule \quad
 \subfloat[MobileNet-v2]
 {\includegraphics[width= 0.3\columnwidth]{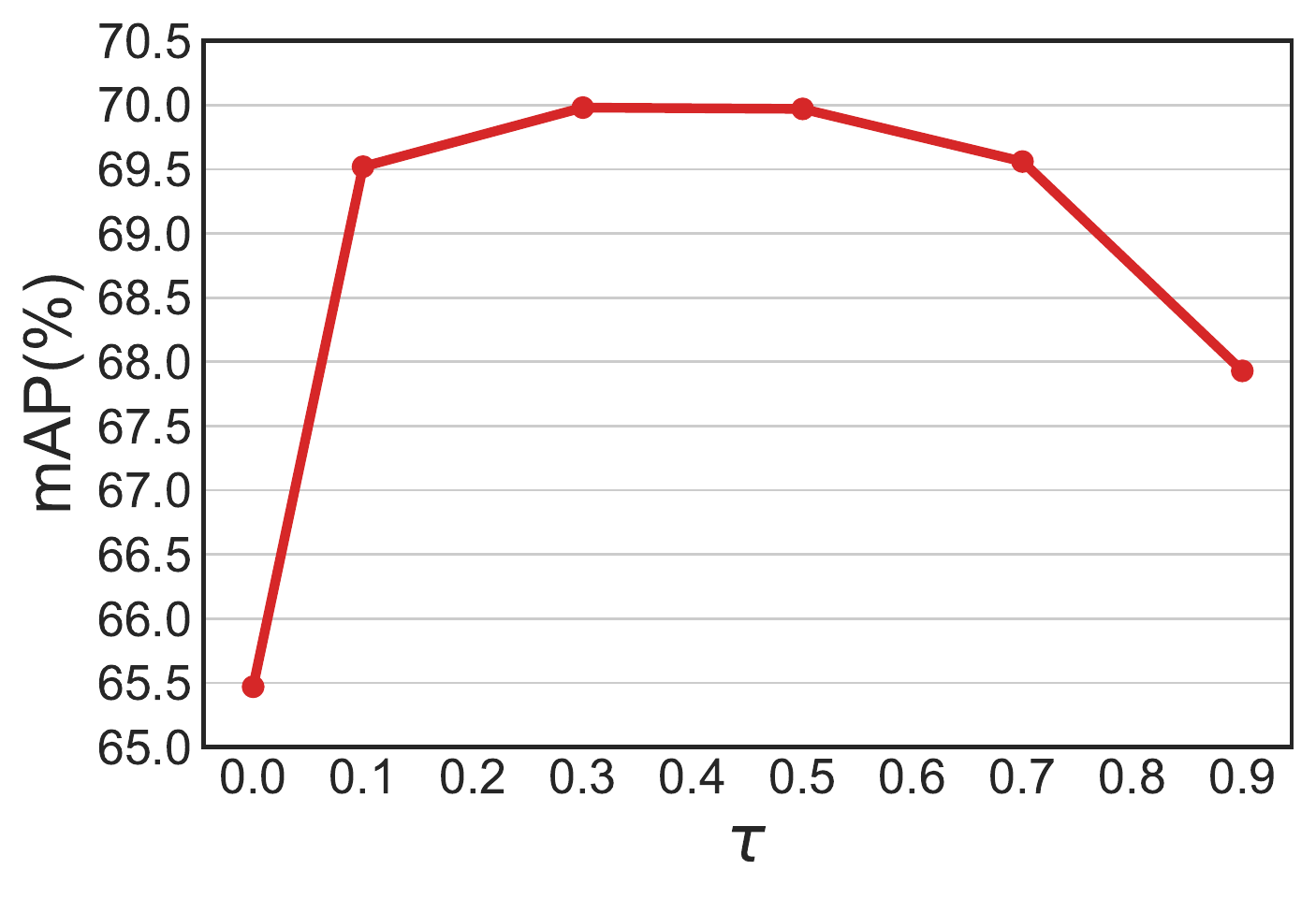}\label{fig:coco-m2thresh}}
 \subfloat[ResNet-50]
 {\includegraphics[width= 0.3\columnwidth]{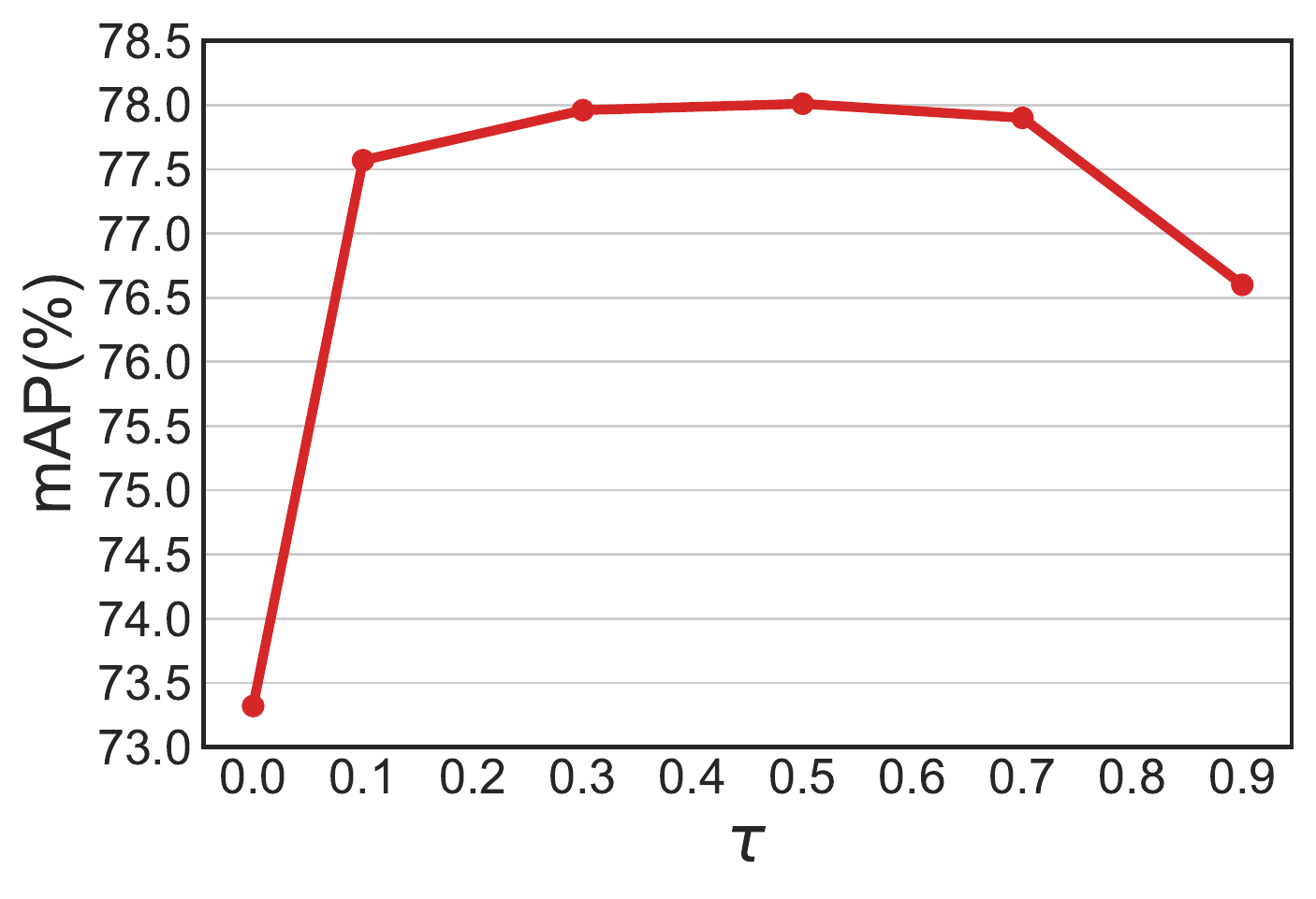}\label{fig:coco-res50thresh}}
 \subfloat[ResNet-101]
 {\includegraphics[width= 0.3\columnwidth]{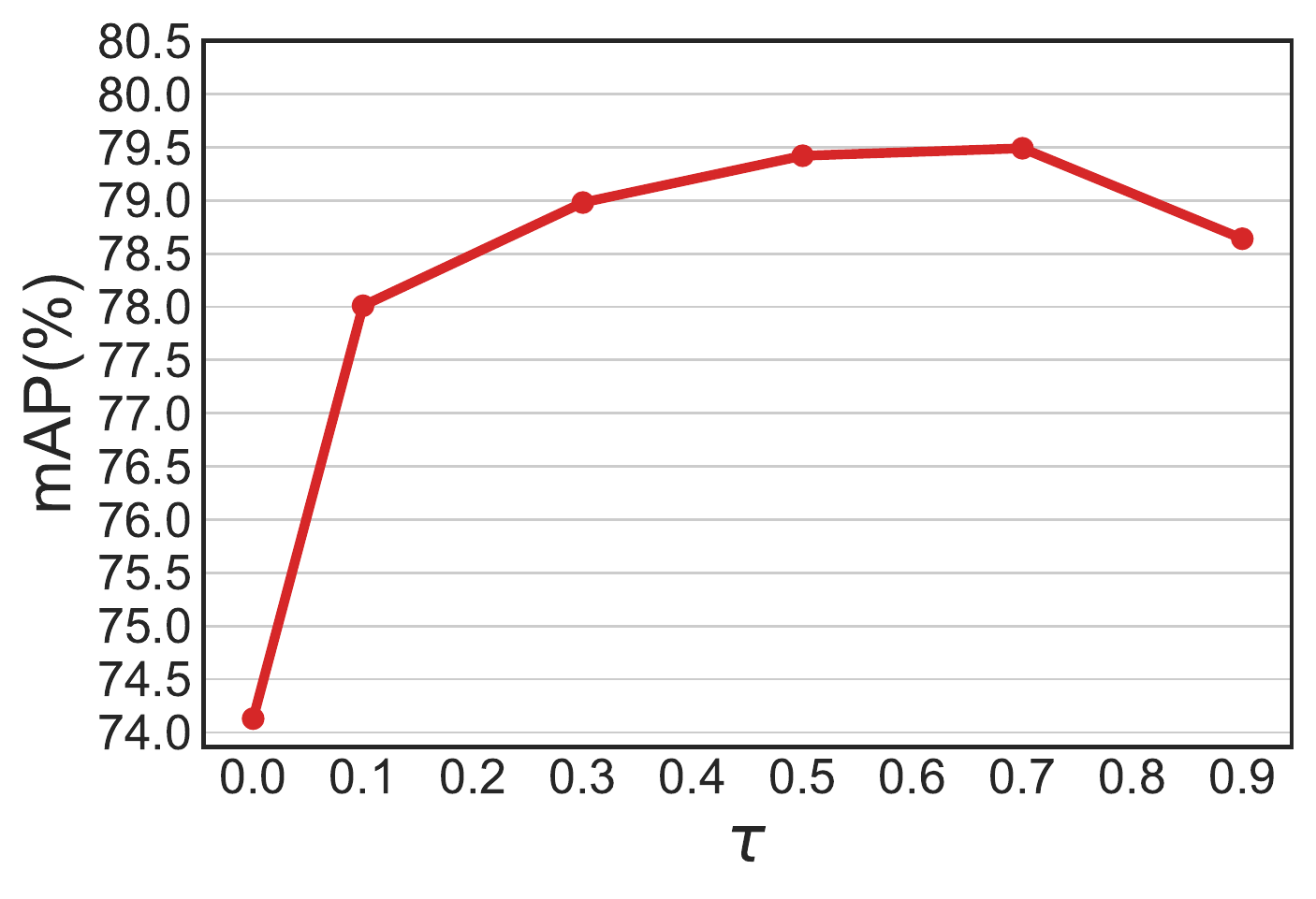}\label{fig:coco-res101thresh}}
\caption{mAP comparisons of our MCAR with different values of  $topN$ and $\tau$. The left three columns are based on PASCAL-VOC 2007 and the right three columns are based on MS-COCO dataset.} \label{fig:hyperpa}
\vspace{-10pt}\label{fig:topntao}
\end{figure*}

\noindent \textbf{Importances of attentional maps selection.}  
In our method, all attentional maps are firstly sorted by global stream score following a descending order and then the $topN$ attentional maps are chosen. In order to further verify the selection strategy, we conduct other criterions to see if the performance is sensitive to the score ranking.  Specifically, we design two criterions to compare with our $topN$ strategy. The first one is that we still sort global stream scores but pick $bottom$ $N$ feature maps. The second one is that we randomly $sample$ $N$ maps among all attentional maps.

For simplicity, we test MCAR with $top4$, $random4$ and $bottom4$ local regions while using the weights trained on MCAR with $top4$ setting in Table~\ref{table:voc-coco-ams}. We can see that the performance of MCAR using high-confidence local regions~($top4$) is significantly better than that of using low-confidence ones~($bottom4$) or random manner~($random4$). This indicates the effectiveness of the local region selection strategy based on the ranking of the global scores.

\begin{table}[t]
	\centering
	\caption{Ablative study of attentional maps selection strategy in MCAR with ResNet-101 backbone and the input size of 448$\times$448.}\label{table:voc-coco-ams}
	\begin{tabular}{|c|c||c||c|}
	\hline
	Methods &Selection criteria   &{VOC 2007} &{MS-COCO}\\
	\hline
	\multirow{3}*{MCAR}  
	 &$bottom4$   &93.8  &81.2\\
	 &$random4$ &94.2  &82.1\\	
	 &$top4$ &\textbf{94.8}  &\textbf{83.8}\\
        \hline
       \end{tabular}
\end{table}

\noindent \textbf{Single or Pair loss?}  
Instead of two-stream learning using a pair of parallel losses in Eq.~\ref{eq:loss}, we design a simple way to train the network utilizing a single loss. Specifically, we firstly fuse global and local prediction scores with a max-wised aggregation 
 \begin{equation}
 \hat {y}^i = \max \big(\hat y_{g}^i , \hat y_{l}^i \big),
 \end{equation}
and then train the network with a single BCE loss as the same in Eq.~\ref{eq:loss},
\begin{equation}
\mathcal {L} = \sum_{i=1}^{M}\sum_{j=1}^{C} y_i^j \log(\hat {y}_i^j) + (1- y_i^j)\log(1-\hat {y}_i^j)\\
 \end{equation}
Using ResNet-101 backbone and keeping the rest settings the same, the experimental results are reported  in Table~\ref{table:voc-coco-loss}. We can see that MCAR with single loss obtains 94.4 mAP on VOC2007 and 82.6 mAP on MS-COCO which improves the baseline by 1.5 and 5.5 mAP, but is 0.4 and 1.2 mAP worse than our main method (with pair loss).

Why is MCAR equipped with a pair of losses better than that of a single loss? The main insight is that global visual processing usually precedes the local one. The pair loss may ensure that after the global stream has been converged fast, then it guides the local stream to find possible local regions. Indeed, we find that the single loss setting usually needs more epochs when arriving at the similar performance. This indicates that the convergence of single loss is slower than that of our pair loss. 
\begin{table}[t]
	\centering
	\caption{Ablative study of learning strategy in MCAR with ResNet-101 backbone and the input size of 448$\times$448.}\label{table:voc-coco-loss}
	\begin{tabular}{|c|c||c||c|}
	\hline
	Methods &Learning Strategy  &{VOC 2007} &{MS-COCO}\\
	\hline
	Baseline                       &single               &92.9  &77.1\\
	\hline
	\multirow{2}*{MCAR}   &single               &94.4  &82.6\\
	                                   &pair &\textbf{94.8}  &\textbf{83.8}\\
        \hline
       \end{tabular}
\end{table}

\noindent \textbf{Number of local regions.} We fix $\tau$ to 0.5 and choose the value $topN$ from a given set $\{0, 1, 2, 4, 6, 8\}$. Note that, $topN=0$ implies we train the model using global stream only, which is equal to our baseline. In the first row of Fig.~\ref{fig:hyperpa}, we show the mAP performance curves when $topN$ is set to different numbers.  First, the mAP performance shows an upward trend with the number of $topN$ gradually being increased. This means that it is useful to improve the multi-label classification performance using more local regions. Second, the performance tends to be stable when $topN$ is set to 4 or 6, which implies that the improvements will be not significant when applying a large $topN$. Third, the performance of a small $topN$,~(\eg, 1, 2, or 4) is significantly better than that of a pure global stream~(\ie, $topN$=0). This further verifies the effectiveness of the proposed selection strategy of generated high-confidence local regions. Another benefit of the region selection strategy is to help reduce the cost of computation resources.

\begin{table*}[t]
	\centering
	\caption{Comparisons of mAP in $\%$ of our methods and baseline on the MS-COCO dataset. Compared to the baseline method, the improvements of our method are highlighted in red.}\label{table:coco-as}
	\footnotesize{
	\begin{tabular}{|c||c|c||c|c||c|c|}
	\hline
	Methods  &\multicolumn{2}{c||}{MobileNet-v2} &\multicolumn{2}{c||}{ResNet-50} &\multicolumn{2}{c|}{ResNet-101}\\  
	\hline
         Input Size                &256  &448   &256  &448   &256  &448 \\     
	\hline\hline
	Baseline                  &61.5 &67.8  &70.1 &75.4  &71.2 &77.1  \\
	\hline
	MCAR~(GAP)        &66.6  {\color{red} $\uparrow$5.1} &74.3  {\color{red} $\uparrow$6.5}  &75.9  {\color{red} $\uparrow$5.8} &78.0 {\color{red} $\uparrow$2.6}  &77.4  {\color{red} $\uparrow$6.2}  &80.5  {\color{red} $\uparrow$3.4}  \\
	MCAR~(GWP)       &69.8  {\color{red} $\uparrow$8.3} &75.0 {\color{red} $\uparrow$7.2}   &78.0  {\color{red} $\uparrow$7.9} &82.1 {\color{red} $\uparrow$6.7}  &79.4  {\color{red} $\uparrow$8.2} &83.8   {\color{red} $\uparrow$6.7} \\
        \hline
       \end{tabular}}
       \end{table*}
\begin{table*}[t]
	\centering
	\caption{Comparisons of mAP in $\%$ of our methods and baseline on the PASCAL VOC 2007 dataset. Compared to the baseline method, the improvements of our method are highlighted in red..}\label{table:voc07-as}
	\footnotesize{
	\begin{tabular}{|c||c|c||c|c||c|c|}
	\hline
	Backbone  &\multicolumn{2}{c||}{MobileNet-v2} &\multicolumn{2}{c||}{ResNet-50} &\multicolumn{2}{c|}{ResNet-101}\\   
	\cline{1-7} 
	Input Size                &256 &448  &256 &448 &256 &448 \\              
	\hline\hline
	Baseline             &85.5           &89.5           &89.1          &91.8  &89.2 &92.9 \\
	\hline
	MCAR~(GAP)    &88.1 {\color{red} $\uparrow$2.6}   &91.3 {\color{red} $\uparrow$1.8} &92.3 {\color{red} $\uparrow$3.2}&94.1 {\color{red} $\uparrow$2.3} &93.0 {\color{red} $\uparrow$3.8} &94.8 {\color{red} $\uparrow$1.9}\\
	MCAR~(GWP)   &88.5 {\color{red} $\uparrow$3.0}   &91.7 {\color{red} $\uparrow$2.2} &92.0 {\color{red} $\uparrow$2.9}&93.7 {\color{red} $\uparrow$1.9} &92.6 {\color{red} $\uparrow$3.4} &94.3 {\color{red} $\uparrow$1.4}\\
        \hline
       \end{tabular}}
\end{table*}

\noindent \textbf{Threshold of localization.} To explore the sensitivity of the $\tau$ in Eq.~\ref{eq:pxpytau}, we fix $topN$ to 4 and test different $\tau$ values from $\{0, 0.1, 0.3, 0.5, 0.7, 0.9\}$. The whole image will be considered as a local region when $\tau$ equals to 0, and it is also equivalent to the  baseline method. We show the mAP performances as the function of $\tau$ in the second row of Fig.~\ref{fig:hyperpa}.
First, we observe that the performance is better when $\tau$ is greater than 0. Second, the performance drops when $\tau$ is either too small or too large. We argue that if $\tau$ is too small, local regions may contain more context information and lack discriminative features because all local regions are close to the original input image.  When $\tau$ is too large, it makes local regions only contain the most discriminative parts of an object and easily leads to over-fitting. It is a good choice when the value $\tau$ is in the interval between 0.3 and 0.7.

\noindent \textbf{Global pooling strategy.}  Encoding spatial feature descriptors to a single vector is a necessary step in state-of-the-art CNNs. The early works,~\eg, AlexNet and VGGNet, use a fully connected layer, and the recent ResNet usually employs global average pooling~(GAP) which outputs the spatial average of each feature map.  Specifically, considering class-agnostic feature map $A$ from the top block of a backbone network. The GAP operation outputs the spatial average of the $A$, returning a vector $\vec f^{a} \in {\mathbf{R}^{d^\prime}}$  with the $k$-th element being
\begin{equation}
f_k^{a} = \frac{1}{h^\prime w^\prime} \sum_{i=1}^{h^\prime}\sum_{j=1}^{w^\prime}  A_{i,j,k}.
\end{equation}  
We denote the output of global maximum pooling~(GMP) as $\vec f^{m}\in {\mathbf{R}^{d^\prime}}$, whose the $k$-th element is 
\begin{equation}
 f_k^{m} = \max \{A_{i,j,k}\}_{i=1~j=1}^{h^\prime~~w^\prime}.
\end{equation} The GMP easily falls into over-fitting because it enforces the network to learn the most discriminative feature. Generally, GAP usually has a better generalization ability than GMP. However, GAP may lead to under-fitting and slow convergence because it equally gives the same importance for all spatial feature descriptors. Our local region localization needs to discover the discriminative region which seems to be opposite to the objective of GAP. In order to alleviate this conflict, we propose a simple solution termed as \emph{Global Weighted Pooling}~(GWP) which is an average of $\vec f^a$ and $\vec f^m$, as 
\begin{equation}
\vec f= \lambda \vec f^a + (1-\lambda)  \vec f^m,
\end{equation}  
where $\lambda \in [0,1]$ is a weight which balances the importance between GAP and GMP. In our paper, the weight $\lambda$ is empirically set to 0.5.

In Table~\ref{table:coco-as}, we can see that MCAR with GWP further boosts performance on MS-COCO dataset. It improves the mAP by 4.1 points and 3.3 points compared to the common GAP on ResNet-50 and ResNet-101 when input size is 448$\times$448.  Nevertheless, the overall performance of GWP is comparable to that of GAP on the PASCAL-VOC dataset as reported in Table~\ref{table:voc07-as}. This may be associated with a specific dataset that the task of PASCAL-VOC is relatively simpler than that of MS-COCO because of small-scale samples, fewer classes and fewer instances per image in PASCAL-VOC. Generally, MCAR equipped with GWP is better than GAP, especially on more challenging tasks.

\noindent \textbf{Network architecture.} The recent state-of-the-art methods usually take ResNet-101 as a backbone to report their performance. However, in real applications, lightweight networks have been widely adopted. To meet such requirements, we extensively evaluate the proposed method with MobileNet-v2 and ResNet-50 besides ResNet-101 on PASCAL-VOC and MS-COCO and report their results in Tables~\ref{table:coco-as} and \ref{table:voc07-as}. The deeper network tends to obtain better performance. This is not surprised because the big network has more parameters and a deeper structure to ensure strong capacity and transferability. Note that our method still has good performance using the lightweight MobileNet-v2. In addition, the proposed method has significant improvements for all backbones. On the MS-COCO dataset, our MCAR with GWP improves the baseline by about 7\% using the input size of 448$\times$448.  

\begin{figure*}
    \captionsetup[subfigure]{labelformat=empty}
    \captionsetup[subfigure]{justification=centering, font=tiny, labelfont=bf}
    \centering
      \captionsetup[subfigure]{oneside,margin={0cm,0cm,5cm,5cm}}
     \subfloat[] [\hspace{-1.4cm} \textbf{Baseline:}  {\it {car, person}} \\ \hspace{-1.04cm} \textbf{MCAR:} {\it {car, person, train}}]{\includegraphics[width= 0.165\textwidth]{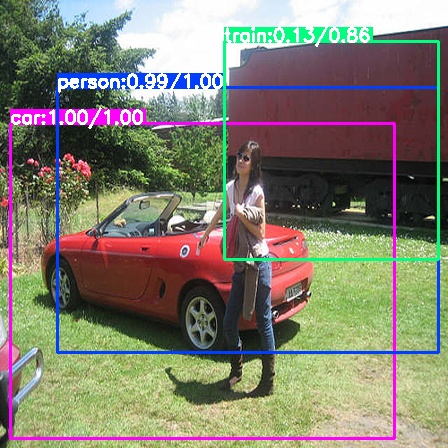}}
    \subfloat[][\it boat\\ boat, person] {\includegraphics[width= 0.165\textwidth]{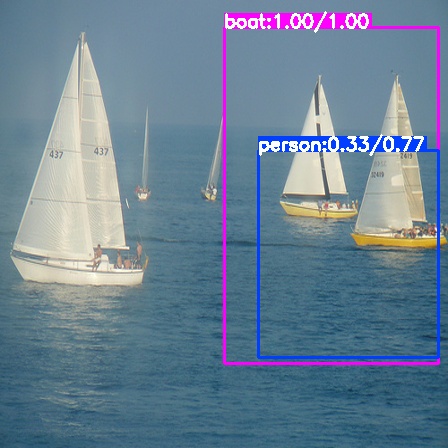}}
    \subfloat[][  \it cat \\ car, cat] {\includegraphics[width= 0.165\textwidth]{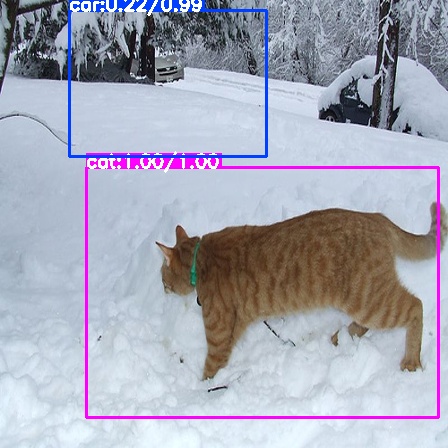}}
    \subfloat[][ \it none \\car] {\includegraphics[width= 0.165\textwidth]{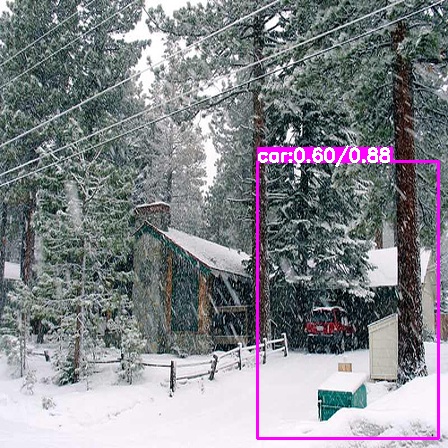}}
    \subfloat[] [ \it car \\ bird, car]{\includegraphics[width= 0.165\textwidth]{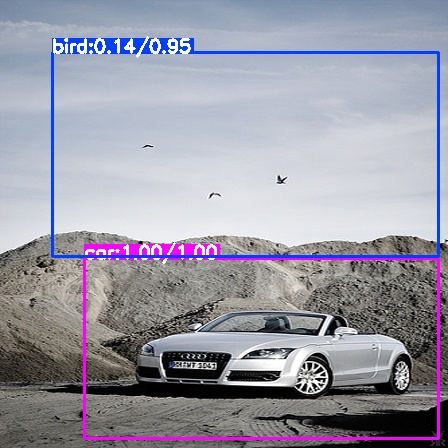}}
    \subfloat[] [ \it horse, person\\horse, person]{\includegraphics[width= 0.165\textwidth]{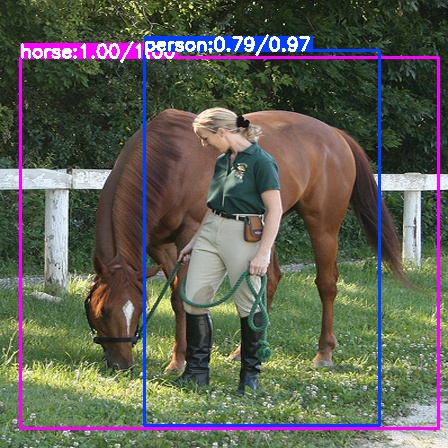}}
    \\ \vspace{-10pt}
     \subfloat[] [\hspace{-1.4cm} \textbf{Baseline:}  {\it {car, person}}\\ \hspace{-1.0cm} \textbf{MCAR:} {\it {car, chair, person}}]{\includegraphics[width= 0.165\textwidth]{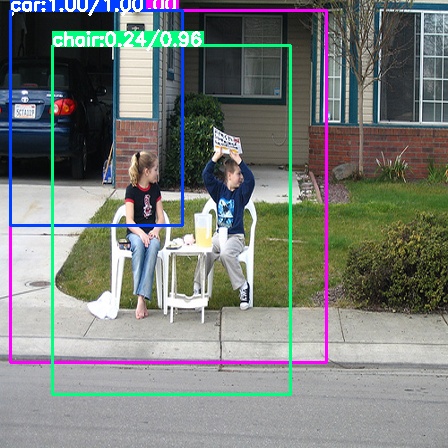}}
      \subfloat[] [ \it mbike, person\\car, chair, mbike, person]{\includegraphics[width= 0.165\textwidth]{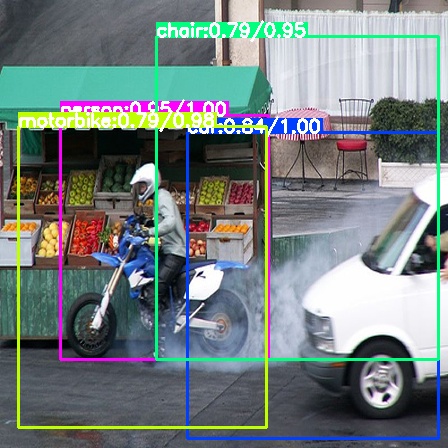}}
      \subfloat[][ \it person \\person, pottedplant] {\includegraphics[width= 0.165\textwidth]{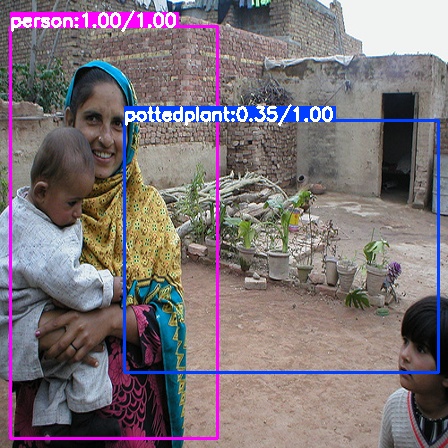}}
      \subfloat[] [ \it person \\cat, person]{\includegraphics[width= 0.165\textwidth]{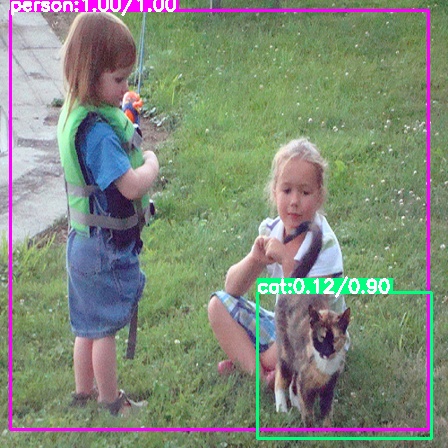}}
      \subfloat[] [ \it none \\ sofa]{\includegraphics[width= 0.165\textwidth]{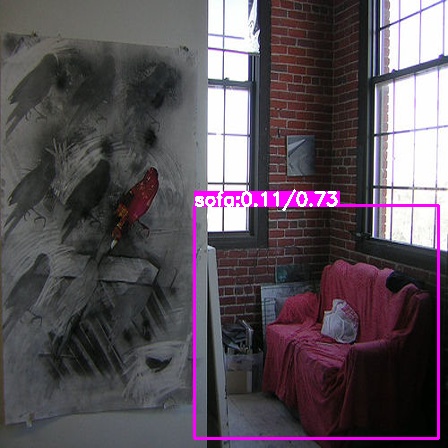}}
      \subfloat[] [ \it bicycle, person\\bicycle, car, person]{\includegraphics[width= 0.165\textwidth]{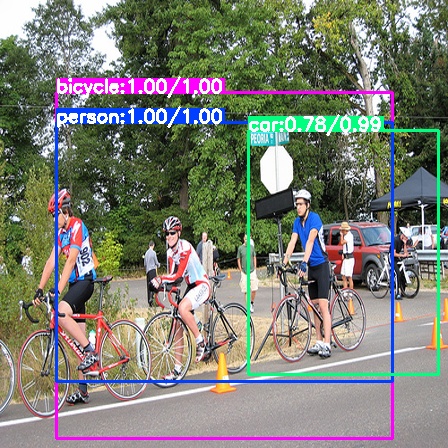}}
    \\ \vspace{-10pt}
    \subfloat[][\hspace{-1.7cm} \textbf{Baseline:} {\it{person}} \\ \hspace{-0.9cm} \textbf{MCAR:} {\it{person, snowboard}} \\\hspace{-1.15cm}\textbf{GT:} {\it{person, snowboard}} ]  {\includegraphics[width= 0.165\textwidth]{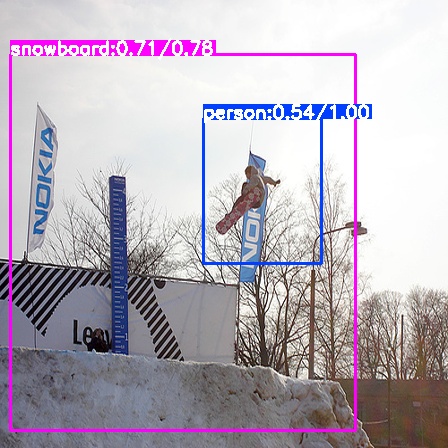}}
    \subfloat[][\it{bench, elephant, person \\bench, elephant, person \\bench, elephant, person}] {\includegraphics[width= 0.165\textwidth]{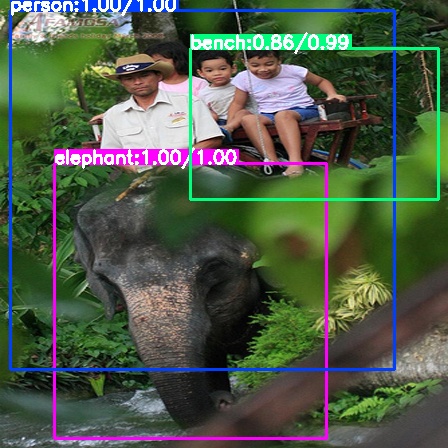}}
     \subfloat[][\it{person, umbrella \\car, person, umbrella\\car, person, umbrella}] {\includegraphics[width= 0.165\textwidth]{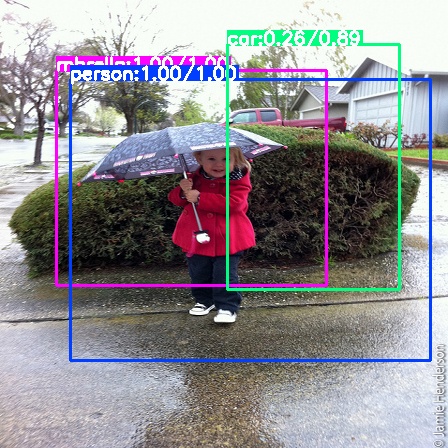}}
     \subfloat[][\it{bbat, bglove, person \\bbat, bglove, bottle, person\\bbat, bglove, bench, bottle, person}] {\includegraphics[width= 0.165\textwidth]{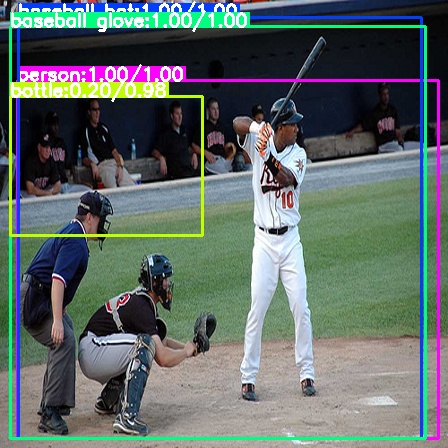}}
    \subfloat[] [\it{bicycle, car, chair, person, truck\\car, kite, mbike, person, tlight, truck\\car, kite, mbike, person, truck}]{\includegraphics[width= 0.165\textwidth]{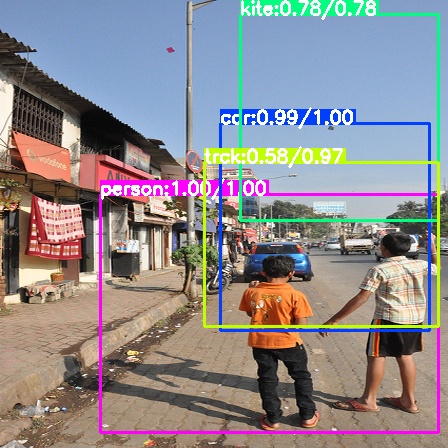}}
         \subfloat[][\it{cow \\ cow, dog \\cow, dog}]{\includegraphics[width= 0.165\textwidth]{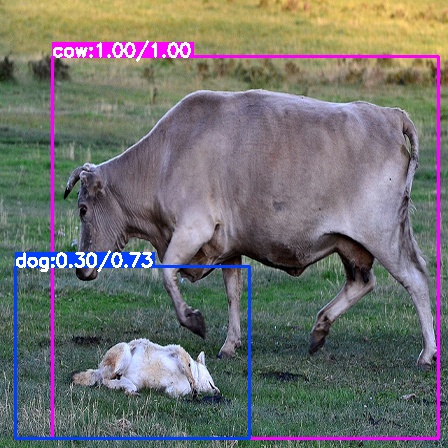}}
   \\ \vspace{-10pt}
    \subfloat[] [\hspace{-1.34cm}\textbf{Baseline:} {\it{cat, person}}\\ \hspace{-0.63cm} \textbf{MCAR:} {\it{cat, cell phone, person}}\\ \hspace{-0.95cm} \textbf{GT:} {\it{cat, cell phone, person}}]{\includegraphics[width= 0.165\textwidth]{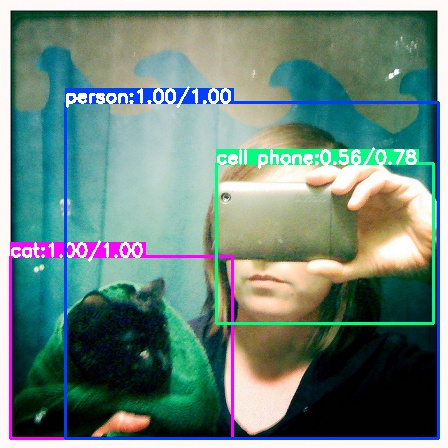}}
     \subfloat[] [\it{airplane \\train \\train}]{\includegraphics[width= 0.165\textwidth]{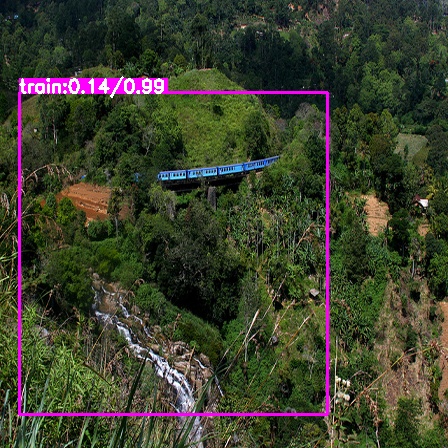}}
         \subfloat[] [\it{clock \\bottle, clock\\ bottle, clock, cup}]{\includegraphics[width= 0.165\textwidth]{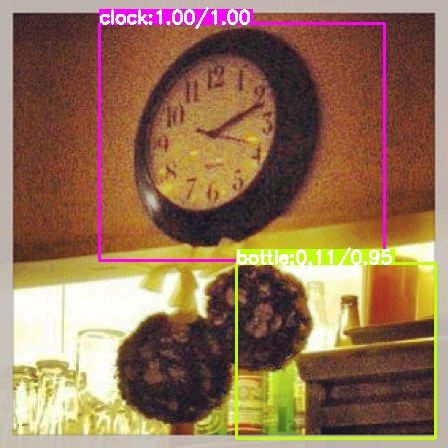}}
         \subfloat[][\it{apple, orange, vase \\ apple, bowl, vase \\ apple, bowl, vase }] {\includegraphics[width= 0.165\textwidth]{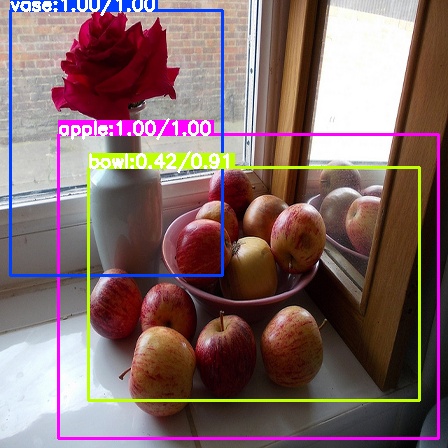}}
          \subfloat[] [\it{cat\\cat, mouse\\cat, couch, mouse}]{\includegraphics[width= 0.165\textwidth]{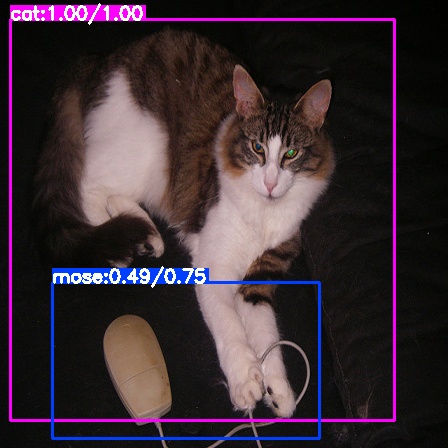}}
         \subfloat[] [\it{laptop, person, tie\\ laptop, person, tie\\book, person, tie}]{\includegraphics[width= 0.165\textwidth]{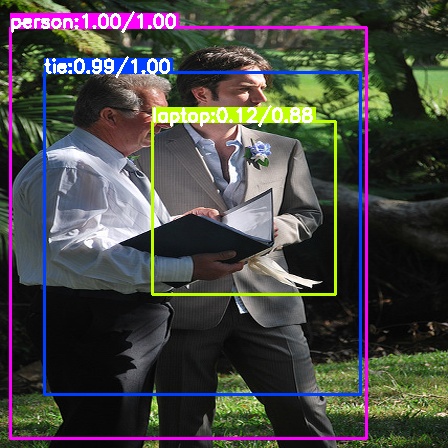}}
 \caption{Selected examples of region localization and classification results on PASCAL VOC 2012 testing images (first two rows) and MS-COCO validation images (last two rows). Our MCAR achieves 94.3\% mAP on the VOC 2012 testing set and 83.8\% mAP on the MS-COCO validation set by using ResNet-101 backbone and the input size of 448$\times$448. Note that these attentional regions are generated by using the model trained on image-level labels only~(without bounding box annotations). Each region box is associated with a category label~($c$), a global stream score~($\hat {y}_{g}^{c}$) and a two-stream score ($\max\{\hat {y}_{g}^{c}$, $\hat {y}_{l}^{c}$\}), organized as ``category name:global score/two-stream score", \eg, ``train: 0.14/0.99" in the image at the fourth row and second column. These region boxes are displayed with conditions on $\hat {y}_{l}^{c} > 0.1$,  $\max\{\hat {y}_l^c, \hat {y}_g^c\} > 0.6$, $topN=4$ and $\tau=0.5$. For each image, one color represents one object category in that image. The proposed two-stream MCAR framework recognizes objects of a wide range of scales, especially for those small-sized or occluded objects, such as the car in (1,3), the bird in (1,5), the cat in (2,4) and the sofa in (2,5) on VOC 2012 testing images and the snowboard in (3,1), the car in (3,3), the dog in (3,6), the cell phone in (4,1), the train in (4,2) and the mouse in (4,5) on MS-COCO validation images, where $(i,j)$ represents the image at $i$-th row and $j$-th column.  It is noteworthy that MCAR may produce incorrect or incomplete predictions when the local region is too small or too blurry such as as the bench in (3,4), the book in (4,6) and the couch in (4,5) on the MS-COCO testing images. (Best view in color and zoom in.)} \label{fig:vis}
\end{figure*}

\noindent \textbf{Input size.} The performance of multi-label recognition is sensitive to the choice of input size. Generally, the larger size tends to get the better performance as reported in Tables~\ref{table:coco-as} and \ref{table:voc07-as} . However, it is more practicable to employ small-sized input on resource-restrict devices. Somewhat surprisingly, MCAR performs better using small inputs. In Table~\ref{table:coco-as} and \ref{table:voc07-as}, we can see that our method always tends to produce more improvements when a smaller input size is employed. This advantage comes from the two-stream architecture which can look at an image in a comprehensive manner (global to local). This indicates that our method is more friendly for low-resolution inputs.

\section{Discussion}\label{discuss}
In this section, we try to understand how the network recognizes multi-objects for a multi-label image via visualizing the produced local regions and discuss why MCAR is a simple and efficient multi-label framework. 

\noindent \textbf{Visualization.} 
To analyze where our model focuses on an image, we show the class-specific attentional regions generated by a multi-class attentional region module in Fig.~\ref{fig:vis}. It can be seen that these attentional regions cover almost all possible objects in each image which is consistent with our initial intention. Furthermore, we can find that global prediction scores of some small-scale objects are low,~\eg~the train in (1,1), the car in (1,3), the bird in (1,5), the chair in (1,1), the cat in (2,4) and the sofa in (2,5) on the PASCAL VOC 2012 testing set and the snowboard in (3,1), the car in (3,3), the dog in (3,6), the cell phone in (4,1), the train in (4,2) and the mouse in (4,5) on the MS-COCO validation images, where $(i,j)$ is the image at $i$-th row and $j$-th column in Fig.~\ref{fig:vis}. This indicates that it is suboptimal to use global image stream solely, especially for small-scaled and partly occluded objects. This limitation would be improved by our two-stream network because it recognizes this type of object from a closer view (high score of two-stream). Compared to the baseline method, our method significantly improves the multi-label image recognition performance. Note that MCAR may produce incorrect or incomplete predictions when local regions are too small or too blurry such as the bench in (3,4), the book in (4,6) and the couch in (4,5) on MS-COCO testing images.

Furthermore, the local region stream is hardly ensured to cover all target objects even if we use a larger number~($topN$). However, the local stream is able to contain a majority of target objects because of the high diversity of local regions. Moreover, our two streams can complement each other by finding missing discriminative regions. Considering this is a weakly supervised problem and the computation efficiency, we think such this situation can be acceptable.

\noindent \textbf{Simplicity.} 
Our framework aims at proposing a simple and efficient method that puts forward to learn global and local image semantics in a single unified model. 
On one hand, we generate object proposals only using the network itself while HCP utilizes external tools such as EdgeBox~\cite{zitnick2014edge} or BING~\cite{cheng2014bing}. On the one hand, our method can efficiently obtain multi-class regions with a parameter-free region localization module because of the parameter share mechanism in Eq.~\ref{eq:linear} and~\ref{eq:cam}. Unlike some existing attention-based methods, they always need a slightly complex module such as LSTM unit in~\cite{jaderberg2015spatial,yu2019delta} or reinforcement learning module in RARL~\cite{chen2018recurrent}.

\begin{table}[t]
	\centering
	\caption{Comparisons of average inference time of per-image between our MCAR (including each component) and baselines with different backbones and input sizes. The time is measured in milliseconds~(ms) on one P40 GPU.}\label{table:inferenetime}
	\resizebox{0.48\textwidth}{!}{ 
	\begin{tabular}{|c||c||r||r||r|r|r|}
	\hline
	\multirow{2}{*} {Methods}  &\multirow{2}{*} {ImgSize} &\multirow{2}{*} {Baseline}   &\multicolumn{4}{c|}{MCAR~($topN$=$4$) }\\  
	\cline{4-7}
	& & &Total  &Global &G-to-L &Local \\
	
	\hline
        \multirow{2}{*} {MobileNet-v2}     &256$\times$256 &6.7 &22.7 &6.6 &9.2 &6.9  \\         
                                                            &448$\times$448 &6.9 &34.3 &6.9 &13.0 &14.4  \\         
	\hline
	   \multirow{2}{*} {ResNet-50}       &256$\times$256 &8.5 &31.7 &8.2 &9.1 &14.3  \\   
	                                                           &448$\times$448 &11.2 &55.4 &11.1 &13.2 &31.2  \\      
	\hline
	 \multirow{2}{*} {ResNet-101}       &256$\times$256 &15.7 &46.8 &16.1 &8.6 &22.1  \\         
                                                             &448$\times$448 &18.4 &82.8 &18.8 &13.5 &50.6  \\     
	\hline
\end{tabular}}
\end{table}

\noindent \textbf{Complexity.} 
The computation complexity linearly grows with the region number (\ie, $topN$). However, it is worth noting that the number of local regions has been significantly reduced by using our framework compared to region-based methods such as HCP (\eg, 500). Our method works well when a small $topN$ (\eg, 4) is used and thus the complexity is controllable and the computation cost is affordable. For example, the number of object proposals in HCP is 500 while we reduce this number to 4. So about 100-time speedup is obtained. 

We test the forward running time of each model using the input sizes of 256$\times$256 and 448$\times$448. This evaluation is conducted on one P40 GPU accelerated by cuDNN v7.4.1. The actual inference time is reported  in Table~\ref{table:inferenetime}. We can see that the total time of our MCAR~($topN$=4) is about 4 to 5 times compared to baselines. This is not surprising because there is at least $topN$+1 times computation cost with our method.  We also report the inference time of each component (global stream, global-to-local and local stream) of our MCAR. It can be seen that those local regions' generation and their forward inference dominate the computation cost of our MCAR, reducing the number of local regions would accelerate inference speed greatly. In addition, our MCAR significantly outperforms the baseline method~(81.9\% vs. 77.1\% mAP on MS-COCO Table~\ref{table:voc-coco}) even if only global image~(without local regions) is taken as input, which implies our method still better than the baseline under the same inference time. Meanwhile, our method needs no additional parameters for generating local regions because of the parameter sharing mechanism between global and local streams.

\section{Conclusion}\label{cons}
We observe that humans recognize multiple objects following two steps. In practice, they usually obey a rule of global to local. Through looking at the whole image at first, people can discover places that need to be focused with more attentions. These attentional regions are then checked closer for a better decision. Inspired by this observation, we develop a two-stream framework to recognize multi-label images from global to local as human's perception system works. In order to localize object regions, we propose an efficient multi-class attentional region module which significantly reduces the number of regions and keeps their diversity. Our method can efficiently and effectively recognize multi-class objects with an affordable computation cost and a parameter-free region localization module. On three prevalent multi-label benchmarks, the proposed method achieves state-of-the-art results. In the future, we will try to integrate the label dependency into our method to further boost the performance. It is also an interesting direction to explore how to extend the proposed method for weakly supervised image detection and semantic segmentation.

\bibliographystyle{IEEEtran}
\bibliography{egbib}
\begin{IEEEbiography}[{\includegraphics[width=1in,height=1.25in,clip,keepaspectratio]{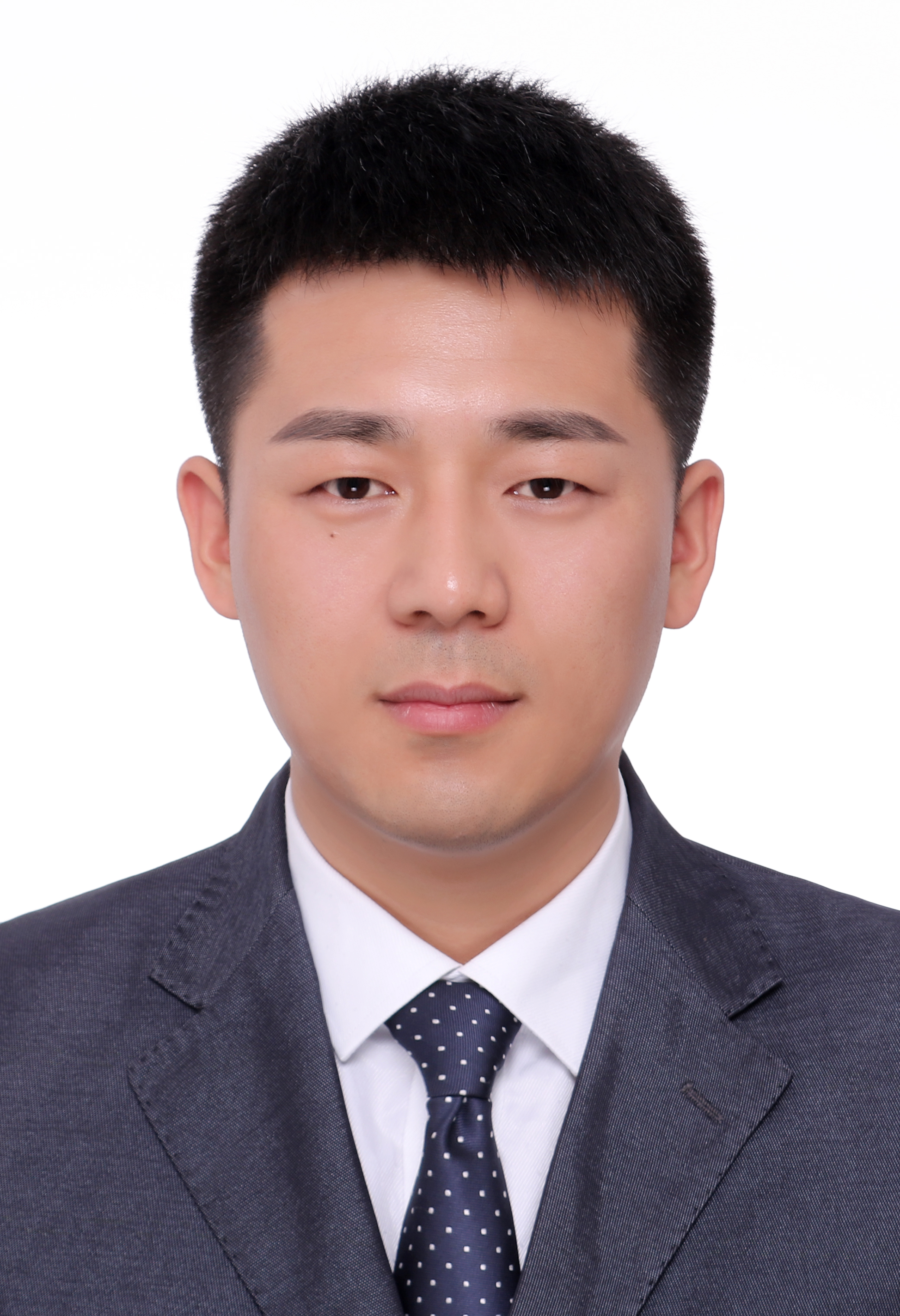}}]{Bin-Bin Gao} is currently a senior researcher in Tencent YouTu Lab. He received the B.S. and M.S. degrees in applied mathematics in 2010 and 2013, respectively, and the Ph.D. degree in computer science from Nanjing University, China in 2018. His research interests are computer vision and machine learning. He has severed as a reviewer for international journals such as the~\emph{IEEE Transactions on Image Processing} (TIP), the~\emph{IEEE Transactions on Neural Networks and Learning Systems} (TNNLS),  the~\emph{IEEE Transactions on Industrial Informatics} (TII), the~\emph{IEEE Transactions on Knowledge and Data Engineering} (TKDE), \emph{Neural Networks}, etc.,  and a program committee member for international conferences such as CVPR, ICCV, ECCV, AAAI, etc.

\end{IEEEbiography}
\begin{IEEEbiography}[{\includegraphics[width=1in,height=1.25in,clip,keepaspectratio]{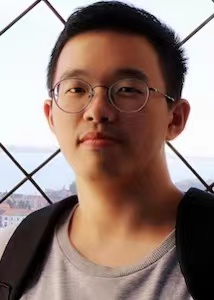}}]{Hong-Yu Zhou} is currently pursuing a Ph.D. degree with the department of computer science, The University of Hong Kong. He received the B.S. degree in Wuhan University, China in 2015, and M.S. degree in the department of computer science and technology, Nanjing University, China in 2018. His research interests include computer vision and machine learning.

\end{IEEEbiography}
\end{document}